\documentclass[pdflatex,sn-mathphys-num]{sn-jnl}% Math and Physical Sciences Numbered Reference Style 
\usepackage[T1]{fontenc}
\usepackage{lmodern}
\usepackage{graphicx}%
\usepackage{multirow}%
\usepackage{amsmath,amssymb,amsfonts}%
\usepackage{amsthm}%
\usepackage{mathrsfs}%
\usepackage[title]{appendix}%
\usepackage{xcolor}%
\usepackage{textcomp}%
\usepackage{manyfoot}%
\usepackage{booktabs}%
\usepackage{algorithm}%
\usepackage{algorithmicx}%
\usepackage{algpseudocode}%
\usepackage{listings}%
\usepackage{placeins} % FloatBarrier and other placing
\usepackage{graphicx} % pictures next to each other
\usepackage{subcaption} % caption for several pictures
\usepackage{setspace} % line spaces
\usepackage{rotating}  
\usepackage{array}     
\usepackage{makecell}  

\usepackage[automake,toc,nonumberlist,nogroupskip]{glossaries} % for abbreviations
\newglossarystyle{mylong}{
 \setglossarystyle{long}
 
 \addtocounter{table}{-1}
 }
 
\makeglossaries

\newacronym{ABEA}{ABEA}{Aspect-based Emotion Analysis}
\newacronym{ABSA}{ABSA}{Aspect-based Sentiment Analysis}
\newacronym{ACD}{ACD}{Aspect Category Detection}
\newacronym{AEC}{AEC}{Aspect Emotion Classification}
\newacronym{AI}{AI}{Artificial Intelligence}
\newacronym{AOI}{AOI}{Area of Interest}
\newacronym{API}{API}{Application Programming Interface}
\newacronym{ASC}{ASC}{Aspect Sentiment Classification}
\newacronym{ATE}{ATE}{Aspect Term Extraction}
\newacronym{BERT}{BERT}{Bidirectional Encoder Representations from Transformers}
\newacronym{CNN}{CNN}{Convolutional Neural Network}
\newacronym{EA}{EA}{Emotion Analysis}
\newacronym{GCN}{GCN}{Graph Convolutional Network}
\newacronym{GHL}{GHL}{Gradient Harmonized Loss Function}
\newacronym{GPT}{GPT}{Generative Pre-Trained Transformer}
\newacronym{GRACE}{GRACE}{Gradient-Harmonized and Cascaded Labeling for Aspect-Based Sentiment Analysis}
\newacronym{HHNN}{HHNN}{Hyperbolic Hopfield Neural Network}
\newacronym{HPO}{HPO}{Hyperparameter Optimization}
\newacronym{IAA}{IAA}{inter-annotator agreement}
\newacronym{KDE}{KDE}{Kernel Density Estimation}
\newacronym{KL}{KL}{Kullback-Leibler divergence}
\newacronym{LDA}{LDA}{Latent Dirichlet Allocation}
\newacronym{LLM}{LLM}{Large Language Model}
\newacronym{ML}{ML}{machine learning}
\newacronym{NLP}{NLP}{Natural Language Processing}
\newacronym{OSM}{OSM}{OpenStreetMap}
\newacronym{OTE}{OTE}{Opinion Term Extraction}
\newacronym{SA}{SA}{Sentiment Analysis}
\newacronym{TN}{TN}{true negatives}
\newacronym{UK}{UK}{United Kingdom}
\newacronym{US}{US}{United States}
\newacronym{VAT}{VAT}{Virtual Adversarial Training}
\newacronym{VLM}{VLM}{Vision Language Model}
\newacronym{VQA}{VQA}{Visual Question Answering}

\begin{document}

\title[Article Title]{EmoGRACE: Aspect-based emotion analysis for social media data}

% LIST OF POTENTIAL JOURNALS:
% Information Processing & Management (https://www.sciencedirect.com/journal/information-processing-and-management)
% Applied Intelligence (https://link.springer.com/journal/10489/aims-and-scope)
% Machine Learning (https://link.springer.com/journal/10994/aims-and-scope)
% Cognitive Computation (https://link.springer.com/journal/12559/aims-and-scope)
% Natural Languate Processing Journal https://www.sciencedirect.com/journal/natural-language-processing-journal
% more generic journals: Information Processing
% Anything from this list? https://ieeexplore.ieee.org/search/searchresult.jsp?queryText=social%20media%20emotion&highlight=true&returnFacets=ALL&returnType=SEARCH&matchPubs=true&refinements=ContentType:Journals
% E.g., IEEE Access https://ieeexplore.ieee.org/xpl/RecentIssue.jsp?punumber=6287639

\author[1]{\fnm{Christina} \sur{Zorenböhmer}}\email{christina.zorenboehmer@plus.ac.at}

\author*[1,2]{\fnm{Sebastian} \sur{Schmidt}}\email{sebastian.schmidt@plus.ac.at}

\author[1,2,3]{\fnm{Bernd} \sur{Resch}}\email{bernd.resch@it-u.at}

% \equalcont{These authors contributed equally to this work.}

\affil[1]{\orgdiv{Department of Geoinformatics - Z\_GIS}, \orgname{University of Salzburg}, \city{Salzburg}, \country{Austria}}

\affil[2]{\orgdiv{Geosocial Artificial Intelligence}, \orgname{Interdisciplinary Transformation University}, \city{Linz},  \country{Austria}}

\affil[3]{\orgdiv{Center for Geographic Analysis}, \orgname{Harvard University}, \city{Cambridge}, \state{MA}, \country{USA}}

%%==================================%%
%% Sample for unstructured abstract %%
%%==================================%%

\abstract{

While sentiment analysis has advanced from sentence to aspect-level, i.e., the identification of concrete terms related to a sentiment, the equivalent field of \acrfull{ABEA} is faced with dataset bottlenecks and the increased complexity of emotion classes in contrast to binary sentiments. This paper addresses these gaps, by generating a first ABEA training dataset, consisting of 2,621~English Tweets, and fine-tuning a BERT-based model for the ABEA sub-tasks of \acrfull{ATE} and \acrfull{AEC}.

The dataset annotation process was based on the hierarchical emotion theory by \citet{Shaver.1987} and made use of group annotation and majority voting strategies to facilitate label consistency. The resulting dataset contained aspect-level emotion labels for \textit{Anger}, \textit{Sadness}, \textit{Happiness}, \textit{Fear}, and a \textit{None} class. Using the new ABEA training dataset, the state-of-the-art ABSA model GRACE by \citet{Luo.2020} was fine-tuned for ABEA. The results reflected a performance plateau at an F1-score of 70.1\% for ATE and 46.9\% for joint ATE and AEC extraction. The limiting factors for model performance were broadly identified as the small training dataset size coupled with the increased task complexity, causing model overfitting and limited abilities to generalize well on new data.
}


%%\pacs[JEL Classification]{D8, H51}
%%\pacs[MSC Classification]{35A01, 65L10, 65L12, 65L20, 65L70}

% =============================
% Centered preprint notice
% =============================
\clearpage  % start a new page
\vspace*{\fill}
\begin{center}
\large
This is a preprint. For the published, peer-reviewed paper see:\\[0.5em]
\textit{Social Network Analysis and Mining} (2026)\\
\href{https://doi.org/10.1007/s13278-026-01585-5}{https://doi.org/10.1007/s13278-026-01585-5}
\end{center}
\vfill
\clearpage  % optional: start next content on a new page

\maketitle

\section{Introduction}

% Sentiment analysis 
One of the largest and most widely researched sub-fields of \gls{NLP} is \gls{SA}. In its most basic form, \gls{SA} assigns sentiment polarity labels (\textit{positive}, \textit{neutral}, and \textit{negative}) to systematically describe the opinions expressed in text \citep{Medhat.2014, Birjali.2021}. Beyond basic polarity classification, \gls{SA} can encompass several other classification tasks, including sarcasm or subjectivity detection \citep{Birjali.2021}. 
\gls{SA} can be categorized into different levels based on the scope of the analysis. Document-level \gls{SA} assigns a single sentiment to an entire document (e.g.,  article), while sentence-level \gls{SA} evaluates individual sentences \citep{Wankhade.2022}. 
% ABSA 
Recent advancements in computing resources, model architectures, and training data availability have allowed \gls{SA} research to progress toward a more fine-grained sub-sentence level, so-called aspect-level analysis \citep{Torfi.2021, Patwardhan.2023, Liu.2020a}. It can generate individual outputs for every word or token, where tokens may even be sub-word units \citep{Mielke.2021}. \gls{ABSA} can involve one or more subtasks, which are either addressed sequentially, i.e., individually and then combined to draw conclusions, or jointly as simultaneous outputs of a model \citep{Luo.2022}. %This comparatively new sub-field, \gls{ABSA}, has received much attention, bringing with it a range of competing taxonomies, state-of-the-art model architectures, and benchmark datasets.

\begin{figure}[h]
\centering
\includegraphics[width=0.9\linewidth]{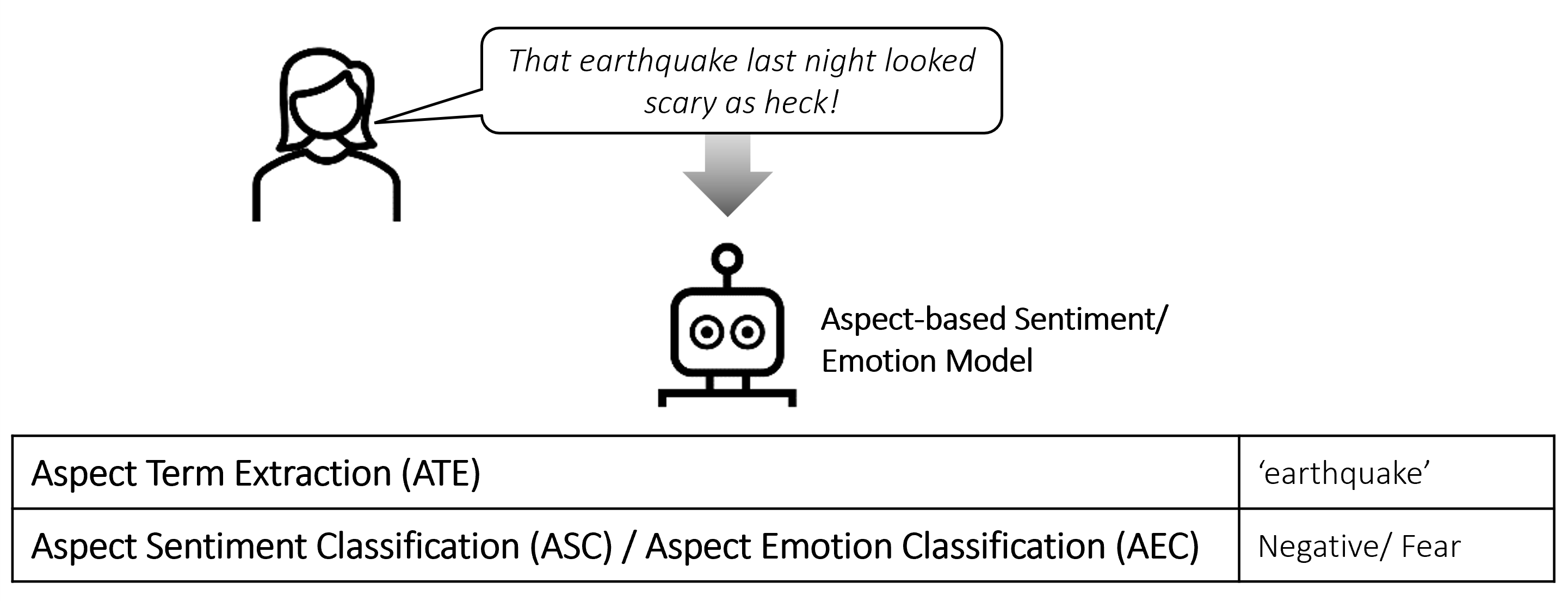}
\caption{Two sub-tasks of ABEA (adapted from \citet{Zhang.2022b}).}
\label{fig:abea_terminology}
\end{figure}

% research gap
\gls{EA} is a sub-discipline of \gls{SA} that categorizes texts according to distinct emotion classes instead of sentiment polarities. In contrast to \gls{SA}, the field of \gls{EA} has yet to undergo a comparable transition to such detailed, sub-sentence level analyses \citep{DeBruyne.2022, Gupta.2018, Mohammad.2021, MatinPour.2021}.
% taxonomy
Given the lack of \gls{ABEA} concepts and taxonomies, we follow \citet{Zhang.2022b} for the definitions of the following two subtasks of \gls{ABEA} (cf. Figure~\ref{fig:abea_terminology}):
1) \gls{ATE}: finding the target word(s) of the sentiment in the text,
2) \gls{AEC}: assigning an emotion label to the respective aspect.
In these definitions, an aspect is understood as the target word(s) of an emotion in a given text. %Based on these sub-tasks, the only difference between \gls{ABSA} and \gls{ABEA} is in the \gls{ASC} task. For \gls{ABEA}, this translates into \gls{AEC} to assign an emotion class to an aspect term.

% social media and ABSA/ABEA
The value of large-scale social media analyses is widely recognized for its direct link to citizen opinions across spatio-temporal dimensions. In \gls{ABSA}, several studies have incorporated social media data, in particular Twitter data. However, no equivalent studies have emerged thus far for \gls{ABEA}, leaving a wide research gap with much potential for new insights, methods, and datasets. This gap is particularly relevant for domains such as disaster management, where social media data is increasingly used for real-time monitoring, situational awareness, and response coordination by gauging public needs and opinions \citep{Yue.2019, Dong.2021, Ma.2014}. 
%Taking a step further, the application of \gls{ABEA} to social media data can offer additional insights by integrating \gls{ABEA} results with the spatio-temporal dimensions of social media data.

% training data generation
%The first research aim tackles a key bottleneck hemming current research on ABEA, namely the lack of publicly available aspect-level emotion datasets.
Even studies reviewing the more prominent field of \gls{ABSA} have highlighted significant dataset-related challenges, such as domain specificity restraining the generalizability of models \citep{Dehbozorgi.2021, Orbach.2021, Luo.2022, Tas.2023, Saeidi.2016}. Filling the training data gap for \gls{ABEA}, particularly for social media data, is a prerequisite for any successful model fine-tuning and application. This paper discusses the logistical challenges of generating such a new \gls{ABEA} training dataset, which encompass the appropriate selection and integration of emotion theory, accounting for human subjectivity \citep{Zad.2021}, and assessing \gls{IAA} at the sub-sentence level.
% implementation of model
Our second research aim is to implement a model capable of \gls{ABEA} through adapting the \gls{GRACE} model by \citet{Luo.2020}. This endeavor presents various sub-challenges, including the choice of model hyperparameters and methods for performance validation.
% research questions
Consequently, we aim to address the following research questions:
\begin{itemize}
 \item \textbf{RQ1}. How can a consistent training dataset for  \acrlong{ABEA} be annotated from social media data? % What methods can be employed to increase annotation consistency for ambiguous annotation tasks, such as aspect-level emotion annotations? How can a measure for inter-annotator agreement be found at the aspect-level?
 \item \textbf{RQ2}. What is the efficacy of repurposing an \acrlong{ABSA} model for \acrlong{ABEA}? Which hyperparameter configurations significantly affect model performance?

\end{itemize}

\section{Related works}

\subsection{Theories of Emotion}

An important basis for \gls{EA} is conceptualizing how emotions are expressed, categorized, and manifested \citep{Acheampong.2020, DeBruyne.2022, Padme.2018, Suciati.2020}. However, there is no clear consensus on this subject matter, making direct comparisons between research outputs challenging \citep{Acheampong.2020, Wang.2020a}.
% emotion models
Originating from the discipline of psychology, two main types of emotion models are commonly used in \gls{EA}: Categorical models, such as \citet{Ekman.1992}, classify emotions into distinct groups, whereas dimensional models, such as \citet{Plutchik.2001}, position them  on continuous scales, such as \textit{happiness-sadness} or \textit{calmness-excitement}. 
% \citep{Ekman.1992} present a categorical model encompassing six basic emotions (\textit{Anger}, \textit{Fear}, \textit{Surprise}, \textit{Sadness}, \textit{Happiness}, \textit{Disgust}), which are universally recognizable, have distinct facial expressions and are present across cultures.  \citep{Plutchik.2001} proposes a dimensional, circumplex model with eight primary emotions (\textit{Joy}, \textit{Trust}, \textit{Fear}, \textit{Surprise}, \textit{Sadness}, \textit{Disgust}, \textit{Anger}, \textit{Anticipation}).
\citet{Shaver.1987} conceptualize a hierarchical emotion graph that incorporates elements of both categorical and dimensional approaches. They identify a set of basic emotions (\textit{Love}, \textit{Joy}, \textit{Anger}, \textit{Sadness}, \textit{Fear}, optionally \textit{Surprise}), which serve as prototypes from which more complex emotions branch out. The closer any two emotion clusters are positioned along the x-axis, the more they are related. Higher y-axis splits indicate stronger sub-branch relationships.
This model acknowledges that emotions are not entirely distinct categories but exist on a continuum, with variations within each basic emotion. The hierarchical emotion model \citep{Shaver.1987} therefore forms the theoretical basis for the \gls{ABEA} model in this study.

\subsection{Training datasets for ABSA \& ABEA}
% training datasets
Alongside fine-tuning methods, an equally defining factor for the adaptation of pre-trained models for downstream tasks is the availability of task-specific training data. 
Current research on \gls{ABSA} is limited by the number of publicly available training datasets and their domain specificity. They are mostly based on customer reviews, which are sourced from e-commerce platforms such as Amazon or Yelp \citep{Do.2019}. The most widely used benchmark datasets “SemEval” \citep{Pontiki.2014, Pontiki.2015, Pontiki.2016a} consist of approximately 3,000~restaurant and laptop reviews and are labeled with aspect terms and sentiment polarity. %Since their publication, \citep{Xu.2020} Xu et al. (2020) expanded the annotations to include opinion terms. 
Alternative \gls{ABSA} training datasets include “SentiHood” \citep{Saeidi.2016}, which contains 5,215~sentences with one or more sentiment target terms related to urban neighborhoods, and “MAMS” \citep{Jiang.2019}, where each sentence includes multiple aspects and sentiments.

% social media analysis
For social media analyses, well-structured datasets like SemEval are unsuitable since they do not resemble the typical complexity, noise, and unstructured nature of social media texts. Three \gls{ABSA} training datasets consisting of Twitter posts can be found in published literature \citep{Mitchell.2013, Dong.2014, Jiang.2011}, which are intentionally biased towards named entities like celebrities, companies, or well-known products. In addition, all predominantly contain neutral sentiment labels. % Of these three, only \citep{Mitchell.2013} and \citep{Dong.2014} are publicly accessible. 
Though further social media \gls{ABSA} training datasets exist, they either do not extract aspect terms \citep{Crowdflower.2015}, or are structured in complex formats like dialogue \citep{Li.2023c}. %An overview of the three relevant datasets is shown in Table 1.

% training data for ABEA
Though various datasets for sentence-level emotions have been created \citep{Acheampong.2020, Mohammad.2021}, \gls{ABEA} research suffers from a lack of publicly available training datasets annotated with both aspect terms and their corresponding emotions \citep{DeBruyne.2022} (cf. Table~\ref{tab:abea_training_datasets}).
Only the datasets in \citet{Maitama.2021, DeGeyndt.2022, Uveges.2022} adhere to the definition of aspect terms as the target of an emotion. Among them, only one dataset is entirely in English, and none are publicly available. Furthermore, none of the datasets share the same emotion classification schema. It should also be noted that all datasets are domain-specific, being related to finance and stock markets \citep{Garcia-Mendez.2023} or Hungarian speeches \citep{Uveges.2022}, making them less applicable and effective for general-purpose fine-tuning of \gls{ABEA} models \citep{Luo.2022}.

\subsection{Models for ABSA}
So far, individual \gls{ABSA} tasks have been more widely researched and have reported higher performance metrics compared to compound tasks \citep{Zhang.2022b}. This performance drop is likely due to the differing nature of the sub-tasks, which are difficult to integrate into a single model \citep{Luo.2020}. For example, \gls{ATE} and \gls{ASC} are inherently different \gls{NLP} tasks, the first being a sequence labeling task and the latter being a classification task. Despite this added complexity, unified models offer several advantages over pipeline methodologies, such as their ability to learn co-dependencies and interactions between the individual tasks \citep{Luo.2020}, while also being able to avoid error propagation \citep{Bie.2021}. %Models solving compound ABSA tasks can be implemented in a fully unified manner, where the model finds all task outputs at the same time, or in a joint manner, where each task is solves sequentially within a single model.
%Both \citet{Wan.2025} and \citet{Wang.2025} propose to use a \gls{GCN} with a position-aware attention mechanism for \gls{ABSA} to integrate syntactic information.

The performance of both single-task and compound-task models is most often assessed against the SemEval datasets \citep{Pontiki.2014, Pontiki.2015, Pontiki.2016a}, with most publications reporting the  macro F1-score as a means for comparison. In this, single-task \gls{ABSA} models generally achieve higher performances than compound-task \gls{ABSA} models. \gls{ATE}-only and \gls{ASC}-only models have reported F1-scores up to 89\% and 91\%, respectively \citep{Wang.2021d, Wang.2021e}. In comparison, compound \gls{ABSA} methodologies lag behind in comparative performance, with pair-extraction models reaching F1-scores in the mid-70\% to 80\%~range \citep{Chen.2020, Liang.2021, Cai.2020}, the highest triplet-extraction models achieving an F1-score of 72\% \citep{Chen.2021a, Xu.2021, Zhang.2021a}, and quad-extraction of 63\% \citep{Bao.2022, Mao.2022, Zhang.2021b}.

% \citep{Cai.2020, Chen.2020, Sun.2019, Liang.2021, Wang.2021d, Wang.2021e, Wu.2020b, Xu.2021a}

\subsection{Models for ABEA}

Many \gls{ABEA} publications do not focus on or even perform \gls{ATE} from the input text. \citet{Padme.2018} perform sentence-level \gls{EA}, first identifying general topics using the topic modeling technique \gls{LDA} and then creating subsets for each topic through keyword filtering. A similar approach is seen in \citet{Suciati.2020}, where aspect terms are predefined categories and their extraction is not part of the methodology. Instead, they compare how well their methods classify emotions for sentences belonging to the predefined aspect categories. Likewise, \citet{Sirisha.2022} report the use of a hybrid ABSA-RoBERTa-LSTM model to classify Tweets into sentiment and emotion classes without specifying how aspect terms were identified. %Instead, they report statistics on sentence-level emotions and sentiments. 
While \citet{Dehbozorgi.2021} seem to extract aspect terms from input text, their method begins by conducting what appears to be sentence-level emotion classification using the \textit{Text2Emotion} Python package.

These examples highlight the fundamentally different task conceptualizations in the small field of \gls{ABEA} research, making systematic method comparisons difficult. A common characteristic emerging from the literature is the widespread use of pipeline methods to address the individual subtasks separately. This shows that \gls{ABEA} methodologies have not yet advanced to joint or unified methods, which are the current research frontier in \gls{ABSA}.

% \citep{Buechel.2017} find that models using four basic emotions (\textit{Joy}, \textit{Anger}, \textit{Sadness}, \textit{Fear}) performed particularly well compared to other emotion constellations. Additionally, \citep{Nandwani.2021} note that these four emotions are among the most commonly used in \gls{EA} and considered to be essential emotion classes, whether they are used on their own or in combination with additional states like disgust or surprise.

\section{Methodology}
\begin{figure}[h]
 \centering
 \includegraphics[width=\linewidth]{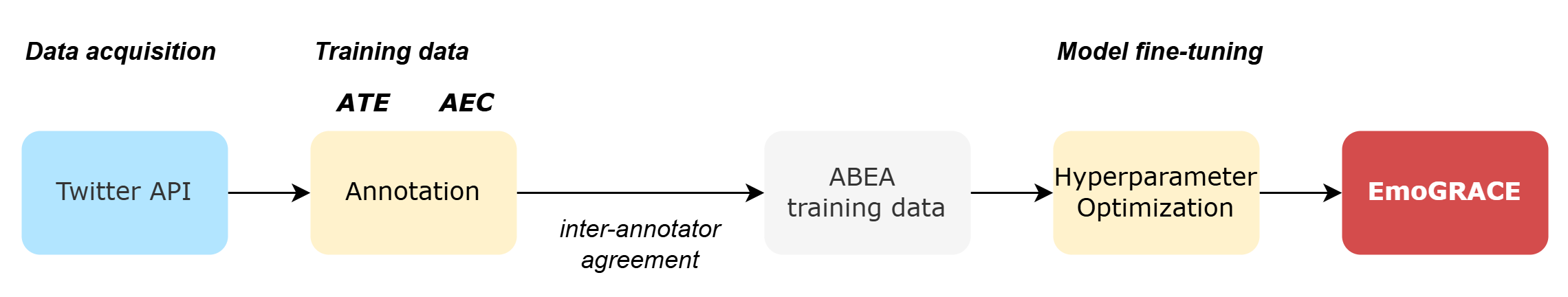}
 \caption{Overview of methodology.}
 \label{fig:workflow}
\end{figure}

The methodology presented in this paper consists of two main steps, namely the creation of an \gls{ABEA} training dataset and the fine-tuning of the \gls{GRACE} model \citep{Luo.2020} for \gls{ABEA}. Figure~\ref{fig:workflow} provides a schematic, high-level overview of our methodology. The code used in this paper is available on \href{https://github.com/Christina1281995/thesis_abea/tree/main}{GitHub}.

\FloatBarrier

\subsection{Data}

Social media data from the microblogging platform Twitter (now X) was used to create a \gls{ABEA} training dataset. A Java-based crawler was employed to collect data, including the Tweet text, a timestamp, and the geolocation, via the v1.1 recent search and streaming \glspl{API} \citep{Havas.2021, Schmidt.2023}. A Tweet’s geolocation was stored either in the form of a point or, more commonly, as a rectangular polygon encompassing a region tagged by the user \citep{Honzak.2024}. %Point locations represented exact coordinates read directly from the user’s GPS device. Alternatively, users can set the location manually. If a user sets a geotag with a large geography, e.g.,  California, the Tweet’s geolocation is stored as a simple rectangular polygon encompassing the respective region. 

Based on the spatial and temporal attributes of the data, an English training dataset consisting of 75\% random and 25\% disaster-related social media posts was assembled. For this, Tweets from the \gls{UK} and the \gls{US} were combined to capture variations in the English language. Tweets for the annotation were then randomly sampled from two larger datasets detailed in Table~\ref{tab:training_datasets}. We incorporated domain-specific training data, assuming that it would contain a higher amount of emotional Tweets and thus addressing the overall class imbalance. The Californian dataset was binary classified with “related” or “unrelated” labels using an XLM-RoBERTa-base model \citep{Barbieri.2022} which was fine-tuned specifically for the classification of disaster-relatedness in social media data \citep{Hanny.2024}. Thus, the training dataset was slightly skewed towards disasters, specifically wildfires and floods, potentially increasing the model sensitivity towards emotion and target detection in natural disaster contexts.

\begin{center}
\begin{table}[ht]
\centering
\caption{Overview of data used to create an \gls{ABEA} training dataset.}
\label{tab:training_datasets}
\begin{tabular}{lm{3.5cm}m{4cm}}
\toprule
\textbf{Attribute} & \textbf{Random Sample} & \textbf{Disaster-Related Sample} \\ 
\midrule
\textbf{Date Range} & 2020-09-01 to 2021-02-28 & 2022-01-01 to 2022-09-30 \\ 
\textbf{Geographic Range} & London, UK & California, USA \\ 
%\textbf{Language} & English & English \\ 
\textbf{Total Dataset Size} & 25,000 & 16,336 \\ 
% \textbf{Random Annotation Subset} & 3,000 & 1,000 \\ 
\textbf{Ratio} & 75\% & 25\% \\ 
\textbf{Bounding Box} & 
\begin{tabular}{@{}c@{}} -126.2721, 31.5850; \\ -112.5449, 42.7965 \end{tabular} &  
\begin{tabular}{@{}c@{}} -0.675965, 51.1500; \\ 0.47999, 51.8297 \end{tabular} \\  
\bottomrule
\end{tabular}
\end{table}

\centering
\end{center}

\FloatBarrier

\subsection{Training dataset annotation}

Due to the lack of publicly available social media \gls{ABEA} training datasets, a new dataset was annotated (cf. Table~\ref{tab:training_datasets}). In total, seven human annotators from the fields of data science and geoinformatics were enlisted for this task. Each Tweet was labeled by three annotators using an online tool configured with Doccano \citep{Nakayama.2018}. The final number of Tweets annotated by three annotators was 2,766. In these Tweets, aspect terms were identified and marked with one of the four basic emotion categories: \textit{Happiness}, \textit{Anger}, \textit{Sadness}, or \textit{Fear}. For dataset consistency, a fifth, null category \textit{None} was also included. The annotation followed an iteratively designed guideline, which set out the tools, rules, and processes for annotations. The guideline was tested on a small pilot dataset of 50~Tweets and revised prior to being used for the annotation. % The full, final guideline is included in Appendix 11.2.

\subsubsection{Emotion detection}

% adapted emotion graph
%The hierarchical emotion graph by \citep{Shaver.1987} was used as the basis for the emotion detection. A well-defined emotion structure for annotation is critical given the many challenges in detecting emotions from text such as contextual shifts in word meanings and implicit emotions \citep{Mohammad.2021, Gupta.2018}.
For the annotations, two amendments were made to the hierarchical emotion graph by \citet{Shaver.1987}: \textit{Love} and \textit{Joy} were joined into one emotion \textit{Happiness}, following the suggestion that even Ekman’s basic emotions can be consolidated due to high similarity between certain pairs (e.g., \textit{Anger} and \textit{Disgust}) \citep{Resch.2016}. Furthermore, \textit{Surprise} was excluded as a stand-alone emotion class as it is a more transient, secondary emotion that triggers another, more dominant one \citep{Ludden.2006}.
The adapted graph was provided to the human annotators to help their decision making process.
% The graph was used as a definitive guide for emotion labeling to increase inter-annotator consistency. By relying on this graph, annotators could make use of the extensive lists of sub-category emotions to help pinpoint the most relevant basic emotion. For example, given a text “I can’t believe I lost everything in the fire” where the decision between anger and sadness is not straightforward, identifying one of the sub-category emotions, such as “anguish” (a sub-category of sadness) can help structure the decision making.

% \begin{figure}[h]
% \centering
% \includegraphics[width=0.9\linewidth]{figures/emotion_hierarchical_graph.PNG}
% \caption{Emotion Annotation Guide, adapted from \citep{Shaver.1987}.}
% \label{fig:emotion_hierarchical_graph}
% \end{figure}

% None class
The inclusion of a neutral class \textit{None} is common practice in \gls{EA} to ensure consistency in the annotations, properly handle texts without apparent emotions \citep{DeGeyndt.2023, Plaza-del-Arco.2024}, and ultimately enhance model robustness. This approach reduces computational complexity in identifying aspect terms and prevents overfitting by providing a label for unemotional texts, mitigating bias toward inapplicable emotion classes.
% Without a \textit{None} class, neutral and unemotional texts would remain unlabeled, which in turn would increase the computational complexity of finding aspect terms in text, since there would not be consistent rules defining aspect terms. The addition of a \textit{None} class also helps prevent overfitting on the four emotions. Without a label for unemotional texts, a model trained on the data may become biased toward assigning one or more of the emotion classes where they are not applicable.

% explicit vs. implicit
Annotators were first guided on how to detect explicit and implicit emotional cues within the text. Explicit emotions are signaled clearly through emotional words, such as "love" in the phrase “I love this time of year”, or "scared" in the phrase “my dogs are scared of the fireworks”. Explicit emotional references typically allow for straightforward identification of the relevant basic emotion. By contrast, implicit emotions require interpretation of the context (e.g., “our government isn't fit to run a race”) or contextual cues like emojis, punctuation (e.g., “...” or “!!!”), or specific word spellings (e.g., “look at thaaaaaat”) to infer the respective emotion.
% unemotional vs. emotional
%Another key aspect of the guideline was the differentiation between unemotional reporting versus emotional accounts of events. 
Additionally, annotators were cautioned that some Tweets may seem emotionally charged but are merely factual reports of events without any intended emotional tone. 
%For example, a statement like "It’s been freakishly dry here" conveys a subjective assessment and therefore carries an emotional tone, whereas a purely factual statement like "Lightning struck a palm tree in Pasadena, causing a fire and power outage" does not. Annotators were asked to be alert in recognizing these distinctions to avoid mislabeling unemotional content as emotional.

% subjectivity
To keep the emotion annotations as objective as possible, the guideline addressed the risk of personal bias. Annotators were warned that their personal opinions, mood, or recent annotation experiences could affect their interpretation of a Tweet's emotional content. For example, if an annotator recently encountered an angry Tweet about wildfires, they might be predisposed to interpret a subsequent Tweet on a related topic as expressing anger, even if the emotion is more subtle or different. To mitigate this, annotators were encouraged to consider the Tweet in various contexts and test different readings before settling on an emotion.

% sarcasm
Lastly, sarcasm was highlighted as a particularly complicated part of emotion detection as it conveys an emotion opposite to the literal meaning of the words. For instance, a sarcastic Tweet like "I lost my key, how wonderful" may literally convey joy but actually express anger or sadness. The annotators were advised to carefully consider the broader context and the potential for sarcasm when determining the true emotional content to ensure the labeled emotion accurately represents the intended sentiment rather than the superficial meaning of the words.

\FloatBarrier
\subsubsection{Aspect term detection}

% aspect terms
Once an annotator identified an emotion in a Tweet, the associated aspect term, i.e., the specific target within the Tweet that the emotion is directed towards, was labeled. In the case of the \textit{None} class, the main subject matter of the Tweet was chosen as the aspect term. If no emotion target or main subject was present, no aspect term was annotated. An example of such a Tweet was “seems fun”.

\begin{center}
\begin{table}[ht]
\centering
\caption{Examples of aspect terms. The respective aspect term is marked in bold.}
\label{tab:aspect_terms_examples}
\begin{tabular}{llp{6.5cm}}
\toprule
\textbf{Category} & \textbf{Type} & \textbf{Examples} \\ 
\midrule
Nouns & Singular noun & “The \textbf{smoke} is getting closer” \newline “my \textbf{car} just died on me.” \\ 
      & Plural noun   & “Just thinking of \textbf{all the holidays} I’ll go on next year.” \\ 
      & Proper noun   & “I miss \textbf{Obama}” \newline “\textbf{Hyde Park} is so good for relaxing after work.” \\ 
\midrule
Pronouns & Subjective personal pronoun & “She lost \textbf{her home}.” \\ 
         & Objective personal pronoun  & “I feel sorry for \textbf{them}” \newline “I love \textbf{you}.” \\ 
         & Demonstrative pronoun       & “\textbf{This} is amazing!” \newline “\textbf{These} are delicious.” \\ 
         & Indefinite pronoun          & “Stay safe \textbf{everyone}” \newline “Well done \textbf{all}!” \\ 
\midrule
Verbs & & “I hate \textbf{voting}” \newline “Would rather be \textbf{sleeping} right now” \\ 
\bottomrule
\end{tabular}
\end{table}

\centering
\end{center}

% difficulty of aspect term identification
The decision to predominantly focus on nouns, pronouns, and verbs stemmed from the otherwise significant increase in complexity in the structure of the labeled dataset. The test pilot revealed that many emotion targets are in fact a highly specific and lengthy combination of entities or actions. For example, in the phrase “the democrats giving these speeches need to get a grip on the facts”, neither “democrats” nor “speeches” fully encapsulate the target of the author’s frustration, requiring the annotation of “democrats giving these speeches” as the aspect term. While this more loosely defined annotation may better reflect the “ground truth” of humanly identifiable emotion targets, it greatly diminishes any systematization of aspect terms, making it exceedingly difficult to set a universal definition. 
%Such complexity hinders systematic information extraction from texts, ultimately impeding intelligible insights. 
Preliminary fine-tuning tests showed that such a loosely defined aspect term definition hinders a model’s ability to learn generalizable \gls{ATE} patterns.
Consequently, the definition of aspect terms was tightened, encouraging the annotators to focus on nouns (singular, plural, or proper), pronouns (personal, demonstrative, or indefinite), or verbs (cf. Table~\ref{tab:aspect_terms_examples}).
The annotators were instructed to be concise and mark only the essential word(s) directly liked to the emotion. %For instance, while adjectives and articles may provide context, they are not considered necessary identifiers of the aspect term itself.

\subsubsection{Inter-annotator agreement}

% joint annotations
Given the complexity of human language, a high degree of ambiguity related to the annotation of aspect terms was expected. % Human interpretation of text may differ considerably and yet be equally valid since there is no definitive correct solution in the absence of the actual author. 
To address the risk of low \gls{IAA} upfront, the annotators were invited to participate in in-person annotation sessions, where they annotated independently but advanced through the dataset at the same pace, allowing for discussions and clarifications at every Tweet. % At the outset of the annotation sessions, the annotators were briefed on the importance that they annotate according to their own best judgement but were invited to discuss and converge with the other annotators in cases of ambiguity. 
This method was inspired by \citet{Luo.2022} and \citet{Maitama.2021}, who employed similar collaborative annotation methods.

% majority vote
All final annotations were determined by means of majority vote %(cf. Figure~\ref{fig:aspect_terms_IAA}) 
\citep{Orbach.2021, Suciati.2020}. At the character level, the annotated aspect terms of all three annotators were compared and only sequences marked by at least two annotators were kept. If all three annotators agreed that no aspect terms were present, the Tweet was included in the final dataset without aspect term labels. In cases where all three annotators marked aspect terms but none of the labeled sequences overlapped, the Tweet was excluded from the final dataset. A similar method was applied for the final emotion labels, with the additional step that cases without a majority emotion (i.e., all different emotion labels) were manually reviewed.
Using this method, 5,704~characters (11\%) from the 51,962~initially annotated aspect term characters were removed from the dataset. %Only 12~instances were left that did not have a dominant emotion. An example of such an instance is: “In the old days the Guardian would be all over a case like this. Nowadays.....nah.", where “Guardian” was annotated by all three annotators (2x sadness, 1x anger) and “Nowadays” was annotated by two annotators (1x sadness, 1x anger). In this Tweet, the first aspect term was automatically labeled with “Sadness”, while the second aspect term was flagged as lacking a dominant emotion. During the post-processing review, the aspect term was evaluated and assigned “Sadness”.

% \begin{figure}[h]
% \centering
% \includegraphics[width=0.9\linewidth]{figures/aspect_terms_IAA.PNG}
% \caption{Visualized example of the majority vote to determine final annotations.}
% \label{fig:aspect_terms_IAA}
% \end{figure}

\FloatBarrier

\subsubsection{Dataset statistics}
\begin{figure}[h]
 \centering
 \includegraphics[width=1\linewidth]{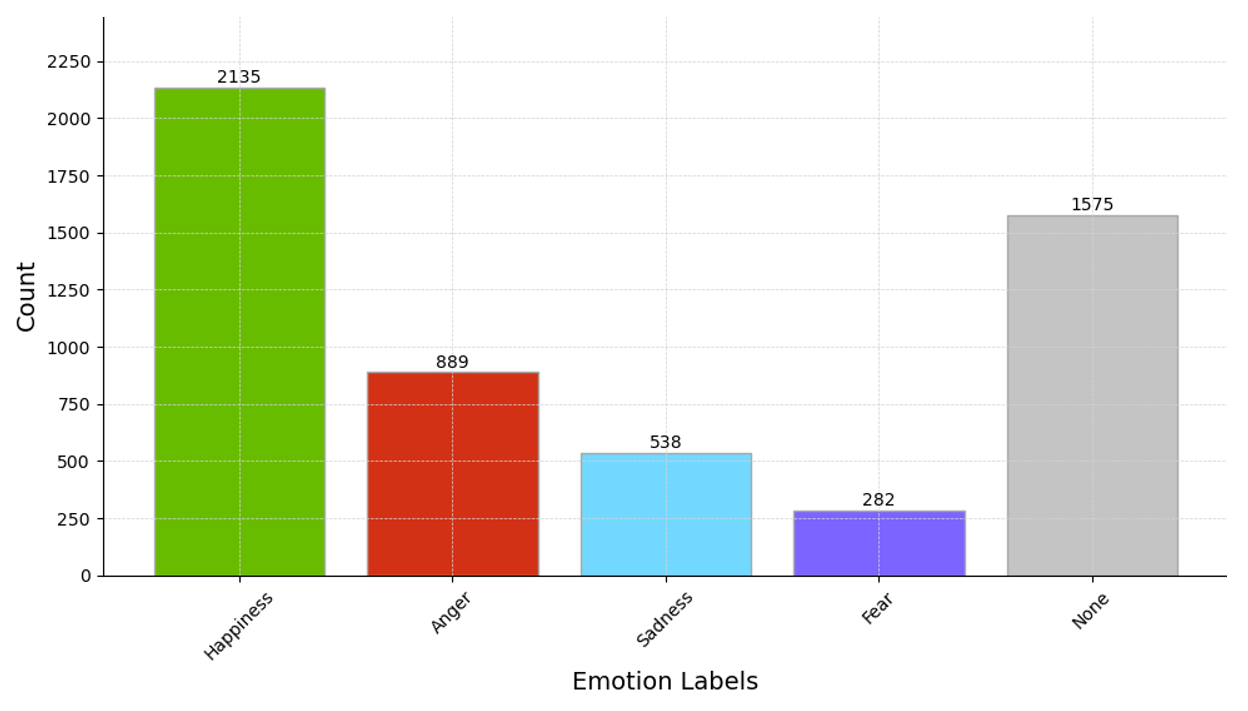}
 \caption{Emotion labels in the ABEA training dataset.}
 \label{fig:emotion_labels}
\end{figure}

% emotion labels
Following the \gls{IAA} assessment, the final \gls{ABEA} training dataset consisted of 2,621~Tweets. The dataset was split into training, validation, and testing subsets using a 70\%, 10\%, and 20\% proportion, respectively. The most common emotion was \textit{Happiness}, which was assigned to 2,153 (39\%) of the aspect terms. By contrast, only 282 (5\%) of all aspect terms were labeled with \textit{Fear}. Figure~\ref{fig:emotion_labels} gives an overview of the label frequencies.

% part of speech tags
A breakdown of aspect terms by part of speech tags revealed that most were nouns. 45\% of all aspect terms were singular nouns such as “fire”, 16\% were singular proper nouns such as “London”, and 13\% were plural nouns such as “days”. Aside from nouns, 6\% of aspect term words were pronouns such as “you”, and 5\% were adjectives such as “relevant” (cf. Figure~\ref{fig:training_data_word_types}). Compared to the \gls{ABSA} datasets by \citet{Mitchell.2013} and \citet{Dong.2014}, this new dataset was both less domain-bound and had more diverse part of speech tags. 
% unique words
While all emotion classes shared some aspect terms (e.g., “you”, “it”, “fire”), there were words unique to each emotion class, such as "shit" for \textit{Anger}, "behavior" for \textit{Fear}, "end" for \textit{Sadness}, "today" for \textit{Happiness}, and "month" for \textit{None}. 
% Words unique to the \textit{Anger} class included “shit”, “business”, and “country". For the \textit{Fear} class, unique words included “behavior”, “cost”, and “conditions”. For \textit{Sadness}, such words were “end”, “face”, and “break”, while the \textit{Happiness} uniquely included “today” and “time”, and the \textit{None} class included “month”, “link” and “lane”.
% tokens
%The new \gls{ABEA} dataset can also be compared with its \gls{ABSA} equivalent, the Twitter dataset by \citep{Mitchell.2013}. Both datasets were tokenized using the \gls{BERT} Tokenizer \citep{Google.2018}, which uses WordPiece tokenization to split text into sub-word segments, to assess the sequence lengths. The token lengths of the Tweets in the new \gls{ABEA} dataset ranged from 1 (e.g., a single emoji) to 70, with an average of 20~tokens. By comparison, the dataset by \citep{Mitchell.2013} contains sequences ranging from 1 to 35~tokens, with an average of 17~tokens. %This indicates that the new \gls{ABEA} dataset has a more varied range of Tweet lengths, with some Tweets consisting of more than one sentence, while the ABSA dataset tends to encompass only one sentence.
% length aspect terms
In the \gls{ABEA} dataset, aspect terms ranged from 2 to 58~characters, with an average of 10. In comparison, aspect terms in the dataset by \citet{Mitchell.2013} ranged from 3 to 84~characters, averaging 11.

\subsection{ABEA model fine-tuning}

\subsubsection{GRACE: Gradient-Harmonized and Cascaded Labeling for Aspect-Based Sentiment Analysis}

With an annotated training dataset in place, a suitable model architecture was needed as a starting point for fine-tuning. In the absence of \gls{ABEA} models, \gls{ABSA} models were considered. Specifically, the search was narrowed down to models that jointly perform the two End-2-End \gls{ABSA} tasks (\gls{ATE} and \gls{ASC}) and were reportedly adapted to social media data.

% applicable ABSA models
The applicable models are shown in Table~\ref{tab:performance_absa_models} and were applied to the Twitter dataset by \citet{Mitchell.2013}. Most models reported considerably higher performance for the individual sub-tasks compared to compound \gls{ATE} and \gls{ASC}. For instance, the \gls{GRACE} model \citep{Luo.2020} achieved 75.7\% on \gls{ATE}, while SpanABSA \citep{Hu.2019} achieved 75.3\% and 75.2\% for \gls{ATE} and \gls{ASC}, respectively. \gls{GRACE} performed best on Twitter data, with a reported F1-score of 58.3\% for joint \gls{ATE} and \gls{ASC} and was therefore selected as starting point for \gls{ABEA} fine-tuning.

\begin{center}
\begin{table}[ht]
\centering
\caption{Performance Comparison of End-2-End ABSA models on social media data.}
\label{tab:performance_absa_models}
\begin{tabular}{m{1cm}m{2cm}m{6cm}m{2cm}}
\toprule
\textbf{Paper} & \textbf{Model} & \textbf{Method} & \textbf{F1-Score [\%]} \\ 
\midrule
\cite{Hu.2019}  & SpanABSA      & BERT-based span-extraction and polarity classification & 57.7 \\ 
\cite{Li.2019a}  & Unified-LSTM  & Unified stacked LSTM with boundary detection and gating & 48.0 \\ 
\cite{Luo.2020}  & GRACE         & Gradient harmonized, cascaded labeling BERT-based model & 58.3 \\ 
\cite{Bie.2021} & MTMVN         & Multitask Multiview network with CNNs and BiGRUs       & 47.9 \\ 
\cite{Ding.2022} & CasNSA        & Cascade BERT-based model with hierarchical structure   & 50.1 \\ 
\cite{Xiang.2023} & BDEP-IAM      & Block-level dependency syntax parsing with interactive attention & 58.0 \\ 
\bottomrule
\end{tabular}
\end{table}
\centering
\end{center}

% GRACE
The basic structure of \gls{GRACE}, which is adapted from the 12-layer, BERT-base model \citep{Devlin.2019}, consists of two computational branches for \gls{ATE} and \gls{ASC} respectively, which share several root layers. %Based on findings from \citep{Jawahar.2019}, the architecturally lowest \gls{BERT} layers capture more basic structures and compositions of the language, whereas the higher layers are more concerned with complex representations. 
\citet{Luo.2020} split \gls{BERT} into two branches, sharing the bottom $l$ (9) layers while reserving the top ~$l:L$ (3) layers for task-specific \gls{ATE} and \gls{ASC} learning.  This architecture enables a shared foundational understanding across both branches. Instead of treating \gls{ASC} as a classification task, both tasks are implemented as sequence labeling to help facilitate co-extraction in a single, unified model. The \gls{ATE} branch labels tokens as $B$ (beginning-), $I$ (inside-), or  $O$ (outside of an aspect term). These labels in turn serve as additional input for the \gls{ASC} branch, where a modified Transformer decoder generates sentiment labels $POS$ (positive), $NEU$ (neutral), $NEG$ (negative), or $O$ (other words). The passing of \gls{ATE} label outputs as \gls{ASC} inputs forms a cascaded labeling approach, which \citet{Luo.2020} argue enhances the model's sentiment judgment.

In addition to cascaded labeling, \citet{Luo.2020} enhance GRACE's training with the use of \gls{GHL} and \gls{VAT}, both of which are designed to improve model robustness when given difficult or imbalanced training data. \gls{GHL}, introduced by \citet{Li.2018a}, mitigates class imbalance by weighting training examples based on their difficulty according to the model's predictions during training. This increases model focus on underrepresented, difficult aspect terms, while also preventing overprediction of easier and more frequent neutral tokens. \gls{VAT}, further enhances model robustness by introducing small adversarial perturbations to the input embeddings during the training process, improving model resilience to data variations \citep{Luo.2020}.

Using these customizations, the fine-tuning was conducted across three training steps, each with a specific number of epochs, learning rate, and setting for \gls{VAT} and \gls{GHL}. The first two training steps focused exclusively on \gls{ATE}. The key difference was that \gls{VAT} was applied in the second step, but not the first. The third step then incorporated the training of the \gls{ASC} branch for the full co-extraction of \gls{ATE} and \gls{ASC}. 
\citet{Luo.2020} configured the hyperparameters based on the SemEval laptop and restaurant datasets. With the extensive customizations for both the model architecture and training process by \citet{Luo.2020}, the \gls{GRACE} model appeared well suited for the complexities of joint \gls{ATE} and \gls{AEC} tasks, learning both tasks in tandem and optimizing the training for robustness.

\subsubsection{Hyperparameter optimization}

%Prior to \gls{HPO}, preliminary testing for \gls{ABEA} fine-tuning revealed several minor errors and deprecated dependencies, which required addressing before functionally conducting model fine-tuning. Specifically, issues related to missing UTF-8 encodings, outdated apex optimizer packages, and indexing errors in the VAT tensor operations were fixed.

The GRACE model was initialized in the same manner as described by \citet{Luo.2020}. Due to the increased classification complexity of emotion classes and considerable structural differences between the SemEval and the new social media \gls{ABEA} dataset, extensive \gls{HPO} was required to identify optimal hyperparameter configurations for \gls{ABEA}.

The \gls{ABEA} dataset was split into train, validation, and test subsets, using the same 70\%-10\%-20\% proportions as \citet{Luo.2020}. In addition to standard hyperparameter categories such as batch size, learning rate, and warm-up proportion, several architectural modifications were tested, including adapting the number of shared layers $l$ for the \gls{ATE} and \gls{ASC} branches, the number of decoder blocks in the \gls{ASC} branch, the classification head architecture, and shallow layer freezing during fine-tuning. 
Across the three overall training steps, the hyperparameters for the learning rate, \gls{GHL}-use, \gls{VAT}-use and batch size were tested individually for each training step. Using the reported optimal \gls{ABSA} hyperparameter configurations as a baseline \citep{Luo.2020}, hyperparameter configurations were adjusted in both directions where possible. % The full and exact hyperparameter configurations are shown in Appendix 9.3 Hyperparameter Settings and Fine-Tuning Results in Table 14.
Table~\ref{tab:results} gives an overview over selected model specifications. 

% adjustment of architectural elements and training settings
The adjustment of architectural elements and training settings enable more influence over the model’s fundamental training and structure. First, \gls{GHL} and \gls{VAT} configurations were adjusted to evaluate their respective impact on performance. Second, the number of shared layers between the \gls{ATE} and \gls{ASC} branches were adjusted in both directions to test the impact of more and less shared learning. When fewer layers are shared, each task-related branch of the \gls{GRACE} model learns more task-specific representations. 
Third, the amount of \gls{ASC} fine-tuning was extended beyond the originally designated epochs of the third training step. Towards this end, fine-tuning iterations were conducted in which \gls{ASC} was trained from the outset, from the second training step, and as an added fourth training step. In the last approach, all other model layers related to \gls{ATE} were frozen to conduct independent emotion learning. 
Fourth, shallow layer freezing was tested by keeping the bottom three, five, and nine~\gls{BERT} layers of the \gls{GRACE} model frozen for the entirety of the fine-tuning process to preserve the originally learned parameters from pre-training. 
Lastly, the model’s single linear classification layer was replaced with more complex classifiers. These included a three-layered multilayer perceptron with rectified linear unit activation functions and a 0.1~dropout rate, as well as three, four, and five layered dense classifiers inspired by the DenseNet structure \citep{Huang.2018a}, which use linear layers and concatenation of intermediate results to preserve input information while improving feature representation.
%While a vast amount of even more complex classification heads would also be technically possible, larger and more complex classifiers risk overfitting, especially on a comparatively small training dataset such as the one used in this thesis and was therefore not considered.

% evaluation of HPO
\gls{HPO} was conducted in a manual grid-search manner, mainly due to the need for code adjustments between iterations and separate validation steps after each fine-tuning iteration. Throughout the \gls{HPO} process, recall, precision and F1 scores were used as performance indicators for both \gls{ATE} and \gls{AEC} task. 
%Recall describes the ratio of true positives (TP) in relation to all actual positives (consisting of TP and false negatives, FN). Precision on the other hand describes the ratio of correct labels (TP) out of all predicted labels (TP and false positives, FP). The F1-score gives a balanced indication of both the precision and recall. 
Accuracy was generally not considered as it includes \gls{TN} in its calculation. At the aspect-level, there is a vast number of \gls{TN} values, i.e., all words in a sentence that are not aspect terms. Their high proportion skews the accuracy metric, which was consistently above 90\%, and does not give a meaningful reflection of model performance.

Using these metrics, the model’s performance was systematically assessed for each \gls{HPO} iteration. A baseline comparison was created with the originally reported hyperparameters by \citet{Luo.2020}. For each hyperparameter category $H_c$ where $c \in~$\{$batch~sizes, learning~rates,...$\}, fine-tuning was performed for each configuration $H_{C_i}$ where $i \in$~\{$config_1, config_2,...config_n$\}. The averaged performance $P(H_{C_i})$ for each configuration was computed from seven sets of macro F1-scores (Equation~\ref{eq:average_performance}). 
Specifically, the F1-score was calculated for both \gls{ATE} and joint \gls{ATE}+\gls{AEC} using the validation data $\mathbb{D}_{\text{val}}$ and after each of the three training steps using the test data $\mathbb{D}_{\text{test}}$ (Equation~\ref{eq:p_train}). After fine-tuning, additional F1-scores were calculated for \gls{ATE} utilizing the external \gls{ABSA} datasets SemEval Laptop $\mathbb{D}_{\text{L}}$ and SemEval Restaurant~2014-16 $\mathbb{D}_{\text{R}}$ \citep{Pontiki.2014, Pontiki.2015, Pontiki.2016a}. For \gls{AEC}, a similar validation was performed using the SemEval Affect~2018 dataset $\mathbb{D}_{\text{A}}$ \citep{Mohammad.2018}.
The performance of each configuration was then assessed against the previous best result. If any configuration from the current hyperparameter category produced a higher average performance than the previously recorded best, that configuration was retained for all subsequent iterations of \gls{HPO}. Otherwise, the configuration was discarded to maintain the original settings across subsequent iterations.

\begin{equation}
\label{eq:average_performance}
P(H_{C_i}) = \frac{1}{7} \big( P_{\text{train}} + P_{\text{extern}})
\end{equation}

\begin{equation}
\label{eq:p_train}
P_{\text{train}} = \sum_{d \in (\mathbb{D}_{\text{val}}, \mathbb{D}_{\text{test}})} \big( f_1^d(ATE, H_{C_i}) + f_1^d(ATE + AEC, H_{C_i}))
\end{equation}

\begin{equation}
P_{\text{extern}} = f_1^{\mathbb{D}_R}(ATE, H_{C_i}) + f_1^{\mathbb{D}_L}(ATE, H_{C_i}) + f_1^{\mathbb{D}_A}(AEC, H_{C_i})
\end{equation}

\vspace{0.4cm}

% validation emotion classification
For validating the \gls{AEC} task, the SemEval Affect~2018 dataset \citep{Mohammad.2018} was used, which consists of sentence-level emotion labels (\textit{joy}, \textit{anger}, \textit{sadness}, \textit{fear}), as no alternative \gls{ABEA} validation datasets exist. When comparing the model's aspect-level emotion inferences with these sentence-level labels, aspect-level emotions were aggregated at the sentence-level, excluding cases with multiple detected emotions per sentence to ensure consistency.
% validation co-extraction
To evaluate the co-extraction of \gls{ATE} and \gls{AEC}, precision, recall, and the F1-score were determined by focusing only on correctly identified aspect terms. A prediction was considered correct only if both the aspect and its emotion were accurately predicted. 
%All metrics were continuously logged along with a performance chart visualization and a bar chart showing the emotion inferences on SemEval Affect 2018.

\section{Results}

%\subsection{Model fine-tuning results for ABEA}
% overview results
The configurations and results for selected hyperparameters, training, and model adjustments are shown in Table~\ref{tab:results}. It provides an overview of the model performance using F1-scores, for the new \gls{ABEA} training dataset and for the validation tests performed on the $\mathbb{D}_L$, $\mathbb{D}_R$, and $\mathbb{D}_A$ datasets. The baseline (configuration~1) is given by the hyperparameters reported by \citet{Luo.2020} as the optimal settings for \gls{ABSA}.

% highest performance
The highest performance - averaged across the \gls{ABEA} validation, \gls{ABEA} test, and SemEval external validation datasets - was achieved with the hyperparameter and model settings in configuration~41, with an overall average of 50.8\%. Configuration~41 achieved slightly higher precision than recall for both the \gls{ATE}-only task and the joint \gls{ATE} and \gls{AEC} tasks (see Table~\ref{tab:metrics_config41}).

\begin{center}
\begin{table}[h]
\centering
\caption{Performance metrics on test data for configuration~41.}
\label{tab:metrics_config41}
\begin{tabular}{lccc}
\toprule
\textbf{Configuration} & \textbf{Recall [\%]} & \textbf{Precision [\%]} & \textbf{F1-Score [\%]} \\ \midrule
ATE Only               & 68.8            & 71.5               & 70.1              \\
ATE + AEC              & 46.1            & 47.7               & 46.9              \\ \bottomrule
\end{tabular}
\end{table}
\centering
\end{center}

\begin{center}

\renewcommand{\arraystretch}{1.2} % Adjust row height for better readability
\setlength{\tabcolsep}{4pt} % Adjust column spacing
\begin{table}[h]
    \centering
    \small % Reduce text size slightly for better fit
    \caption{Hyperparameter Settings and Results. 
    The table shows a selection of hyperparameter configurations with settings that were propagated onwards. Empty cells imply original settings as shown in Config.~1}
    \label{tab:results}
        \begin{tabular}{l|c|c|c|c|c|c}
        \toprule
        \multicolumn{1}{l}{\textbf{Config. Nr.}} & 1 & 6 & 12 & 14 & 33 & 41  \\
        \midrule
        \multicolumn{1}{l}{\textbf{Categories}} & Orig Settings & Batch Size & Epochs & Warmup & AEC Layers & VAT  \\
        \midrule
        
        \multicolumn{7}{c}{\textbf{Hyperparameter Settings}} \\
        \midrule
        Dropout & 0.1 &  &  &  &  &     \\
        Weight Decay & 0.01 &  &  &  &  &     \\
        Batch Size & 32 & 8 &  &  &  &    \\
        Gradient Acc. & 2 &  &  &  &  &     \\
        Epochs Step 1 &  5&  & 10 &  &  &    \\
        Epochs Step 2 &  3&  & 9 &  &  &     \\
        Epochs Step 3 &  10&  & 25 &  &  &     \\
        Warmup Method & Linear &  &  & Constant &  &     \\
        Warmup Proportion & 0.1 &  &  &  &  &     \\
        LR Training Step 1 & 3.0E-05 &  &  &  &  &     \\
        LR Training Step 2 & 1.0E-05 &  &  &  &  &     \\
        LR Training Step 3 & 3.0E-06 &  &  &  &  &     \\
        Shared Layers &  9&  &  &  & 5 &     \\
        Nr. of AEC Layers &  2&  &  &  & 6 &     \\
        % GHL &  &  &  &  &  &  &   \\
        VAT &  enabled in 2nd step&  &  &  &  & disabled  \\
        % Frozen Layers &  &  &  &  &  &  &   \\
        % Classification Head &  &  &  &  &  &  &   \\
        \midrule
        
        \multicolumn{7}{c}{\textbf{F1 for the New ABEA Dataset [\%]}} \\
        \midrule
        ATE Validation & 65.3 & 67.5 & 67.8& 68.5& 68.6& 70.1\\ 
        ATE + AEC Validation & 42.0 & 41.4& 42.5& 43.6& 44.7& 46.9\\ 
        ATE Test & 65.9 & 67.0& 68.2& 67.2& 66.8& 67.1\\ 
        ATE + AEC Test & 42.9 & 45.1& 44.7& 43.6& 45.9& 47.1\\ 
        ATE on \(D_R\) & 63.4 & 63.2& 64.0& 64.5& 63.5& 64.3\\ 
        ATE on \(D_L\) & 46.5 & 46.1& 46.8& 47.4& 45.8& 46.2\\ 
        AEC on \(D_A\) & 10.5 & 12.5& 13.1& 13.2& 13.9& 14.1\\ 
        \textbf{Average} & \textbf{48.1} & \textbf{49.0}& \textbf{49.6}& \textbf{49.8}& \textbf{46.5} & \textbf{50.8}\\ 
        \bottomrule
    \end{tabular}
\end{table}

\centering
\end{center}

% handling overfitting
Configurations~2–5, which adjust dropout and weight decay, as well as configurations~47-49, which implement layer freezing during training, were designed to address the likely tendency for overfitting of a complex \gls{BERT}-based model when fine-tuned on a comparatively small dataset. During configurations~2–5, the initial low number of training epochs was maintained, while dropout was increased across all \gls{BERT} layers in both task branches and the weight decay parameter in the loss function was increased to penalize large weights. The results, however, showed that in particular the performance on the SemEval validation datasets dropped to lower levels than those of the initial configurations. For configurations~47–49, which applied shallow layer freezing, no improvements were recorded for either the \gls{ABEA} dataset metrics or the SemEval validation metrics.

% classification heads
Configurations~50–54, in which the classification heads were exchanged with slightly more complex architectures, exhibited the most prominent trade-offs between gains in one metric and declines in others. All configurations, including the baseline, consistently demonstrated low performance in the emotion validation on $\mathbb{D}_A$.

% hyperparameter adjustments
Hyperparameter adjustments that were ultimately kept included a reduction of the batch size (configuration~6), an increase in the number of training epochs (configuration~12), an adjustment to the warm-up method (configuration~14), a reduction in the number of shared layers between the \gls{ATE} and \gls{AEC} branches and an increase of \gls{AEC} branch layers (configuration~33), and the omission of \gls{VAT} in the second training step.

\begin{figure}[h]
 \centering
 \begin{subfigure}{0.49\textwidth}
 \centering
 \includegraphics[width=\textwidth]{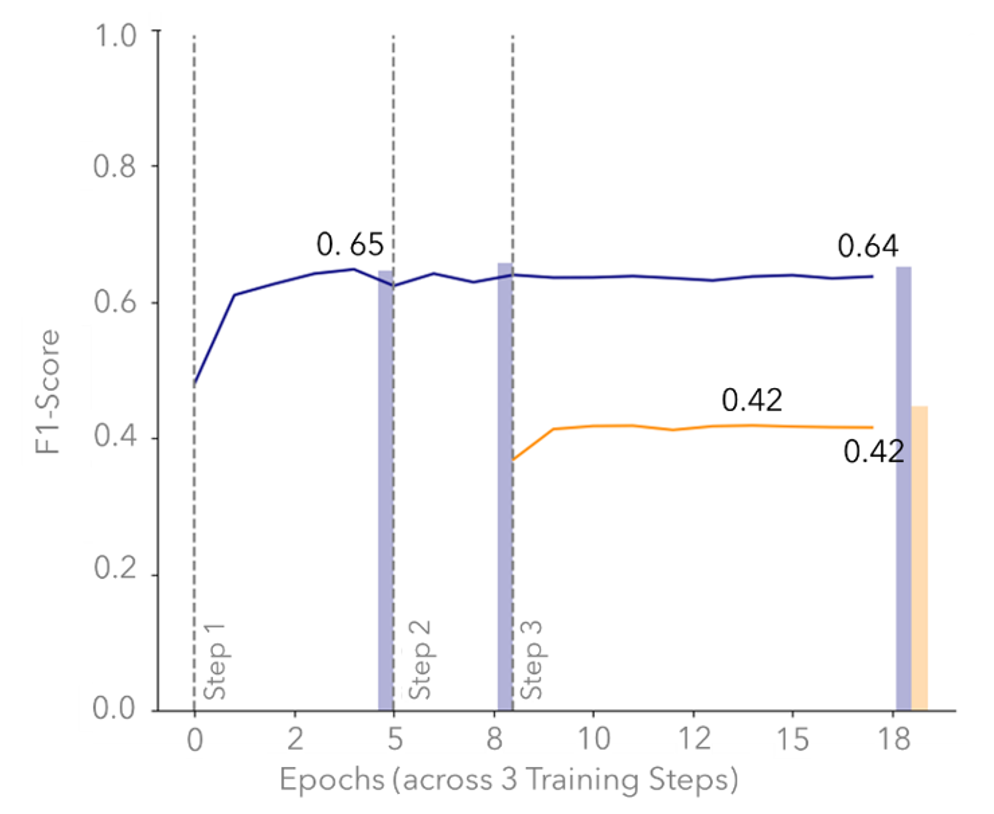}
 \caption{F1-score per Epoch for Configuration 1.}
 \label{fig:progression_model_performance_config1}
 \end{subfigure}
 \hfill
 \begin{subfigure}{0.49\textwidth}
 \centering
 \includegraphics[width=\textwidth]{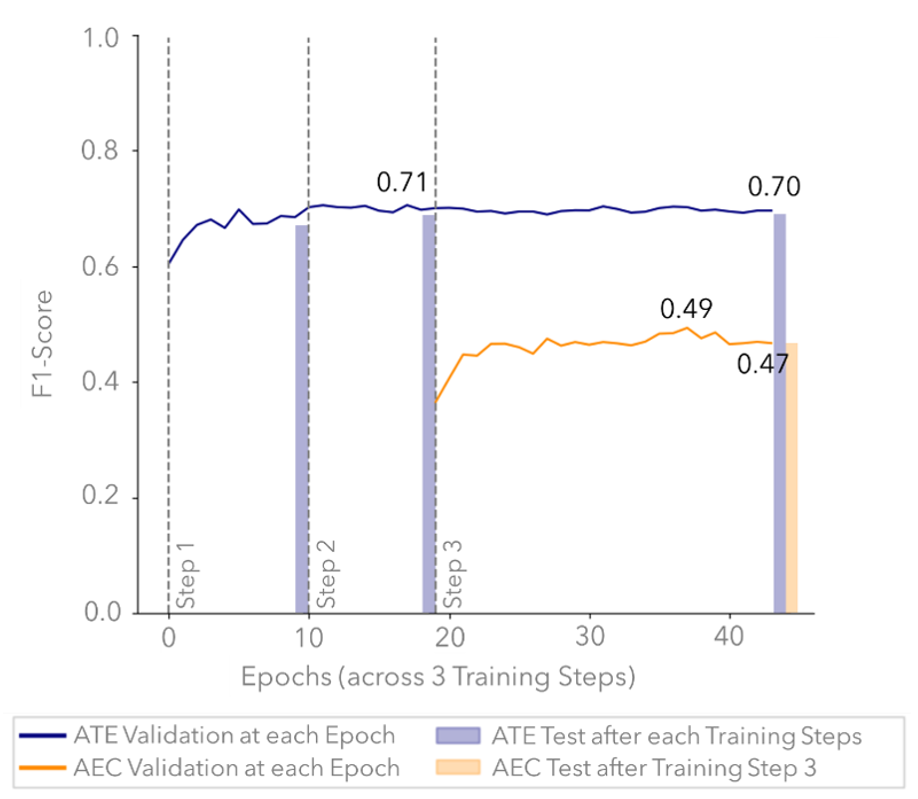}
 \caption{F1-score per Epoch for Configuration 41.}
 \label{fig:progression_model_performance_config41}
 \end{subfigure}
 \caption{Progression of model performance during 10-fold cross validation.}
 \label{fig:progression_model_performance}
\end{figure}

% cross validation
The fine-tuning results for the baseline (configuration~1) and the highest average configuration (configuration~41) were further validated through 10-fold cross-validation. The results are displayed in Table~\ref{tab:cross_validation_results} and show a 6-point improvement for the joint \gls{ATE} and \gls{AEC} tasks over the baseline. 
Figure~\ref{fig:progression_model_performance} shows the progression of model performance during the fine-tuning process on the first of the ten folds for configuration~1 and 41. In both instances, model learning for \gls{ATE} plateaued before the third training step. In the third training step, \gls{AEC} fine-tuning was performed in conjunction with \gls{ATE} fine-tuning. With the hyperparameter settings in configuration~1, model learning for \gls{AEC} remained minimal, starting at an F1-score of 38 and peaking at 42. With the settings in configuration~41, the model progressed from 38 to 49.

\begin{center}
\begin{table}[ht]
\centering
\caption{10-Fold Cross Validation for Configuration~1 (Original Settings by \cite{Luo.2020}) and 41 (Highest Performance). The configuration with the highest performance is shown in bold.}
\label{tab:cross_validation_results}
\begin{tabular}{@{}llllllllllll@{}}
\toprule
\multicolumn{12}{c}{\textbf{Configuration 1}} \\ \midrule
\textbf{Cross Fold} & 1    & 2    & 3    & 4    & 5    & 6    & 7    & 8    & 9    & 10   & Average \\ \midrule
\textbf{ATE}        & 63.8 & 64.0 & 65.7 & \textbf{69.3} & 67.0 & 65.3 & 65.3 & 66.2 & 64.4 & 67.4 & 65.8          \\
\textbf{ATE+AEC}    & 41.8 & 38.4 & 41.9 & 43.1 & 44.3 & \textbf{46.4} & 40.6 & 41.5 & 40.1 & 43.0 & 42.1          \\ \midrule
\multicolumn{12}{c}{\textbf{Configuration 41}} \\ \midrule
% \textbf{Cross Fold} & 1    & 2    & 3    & 4    & 5    & 6    & 7    & 8    & 9    & 10   & \textbf{AVG.} \\ \midrule
\textbf{ATE}        & 70.3 & 67.9 & 71.2 & \textbf{71.5} & 69.0 & 64.5 & 71.4 & 69.8 & 63.6 & 70.9 & 69.1          \\
\textbf{ATE+AEC}    & 49.3 & 45.9 & \textbf{52.3} & 49.6 & 46.7 & 47.7 & 47.2 & 47.7 & 44.1 & 49.0 & 48.0          \\ \bottomrule
\end{tabular}
\end{table}
\centering
\end{center}

\FloatBarrier

\section{Discussion}

\subsection{Training Dataset Annotation}
The labeled \gls{ABEA} dataset was created by human annotators through a process that aimed to balance strict, well-defined rules with the flexibility needed for the naturally ambiguous interpretation of emotions. Differences between theoretical conceptualization and practical applicability of the emotion classes and aspect terms warrant further discussion, as do the dataset results regarding their suitability for \gls{NLP} model training.

\subsubsection{Emotion ambiguity}
% complexity of emotions
The inherent subjectivity of human emotions, combined with the lack of contextual information such as facial expressions, spoken tone, or images, makes aspect-level labeling of emotions in written text a considerable challenge. Throughout the annotation process, many Tweets were encountered that could reasonably be labeled with two or more emotions, depending on the Tweet’s unknown context, to which none of the annotators had access. 
%An example of such a Tweet is “Politics is all over the place at the moment”, which could contain both \textit{Sadness} or \textit{Anger}.
%In cases with explicit and clear ques in the text, the identification of emotions was comparatively straightforward, as with the detection of \textit{Happiness} in “Lovely day for diving! \#divemaster \#scubadiving”. In cases of implicit emotions, much more room was left for an annotator’s own interpretation and potential misunderstanding.
%Especially for a dataset intended for \gls{NLP} model training, annotation consistency was therefore considered a topmost priority. 
% struggles
Despite the clear and logically sound structure of the emotion graph proposed by \citet{Shaver.1987}, annotators repeatedly struggled to make definitive decisions on the emotion labels when encountering instances of sarcasm, rhetorical questions, or hypothetical statements. 
% sarcasm
In many cases, sarcasm is self-evident to humans when provided with visual and auditory context. Even in writing, sarcasm can often be detected based through exaggerated emotional expression. Aside from such clear-cut cases, some Tweets fell into the fuzzy boundaries between earnest emotion and sarcasm, posing a challenge for the annotators to confidently label both the emotion and the aspect term.
%An example was “Taxes to fund fire departments are unfair to people whose houses didn’t catch on fire.”, which could either be sarcasm indicating annoyance with anyone holding that opinion, possibly the “people whose houses don’t catch fire”, or if genuine, it may convey anger or sadness directed at “taxes”.
% hypothetical scenarios
Similarly challenging were hypothetical scenarios, such as “if the Lakers don’t win this game, I might smash something fr”. These types of statements often involved conditional structures that juxtaposed different possible outcomes, each tied to a different emotion. %In the Lakers game example, annotators debated whether the author was already disappointed or angry due to the potential game loss, neutrally reporting future disappointment at the potential game loss, or in a state of nervous anticipation. In this case, anxiousness, belonging to the basic emotion ‘Fear’ was ultimately determined as the most likely emotion.
Among other persistent challenges in distinguishing emotions, distinguishing between caring (\textit{Happiness}) and worry (\textit{Fear}), or frustration (\textit{Anger}) and disappointment (\textit{Sadness}) was frequently debated as conceptually overlapping emotions. The group annotation system and majority vote evaluation helped establish a consensus on these ambiguous cases.

% future developments
Following the annotation process, the necessity of explicitly accounting for sarcasm and hypothetical statements stands out as a critical area for improvement. Future annotation guidelines may benefit from introducing distinct categories for these occurrences or excluding them from the dataset altogether. By addressing these edge cases directly, annotators could avoid ambiguity in their labels, ensuring greater consistency and improving the overall quality of the training data for emotion recognition models.

\subsubsection{\textit{None} Emotion Class}

To ensure model consistency and reduce fine-tuning bias, a \textit{None} class was added to account for neutral, unemotional statements. 
From an emotion labeling perspective, assigning \textit{None} to an aspect term is counterintuitive, since an aspect term is by definition the explicit target of an emotion in a text. The class was nonetheless used to capture instances of neutral statements, without which an \gls{NLP} model might force unemotional texts into one of the emotion classes, leading to bias and misrepresentations.

During initial experiments using the original \gls{GRACE} hyperparameter settings, removing the \textit{None} class led to a nearly 10~percentage-point decrease in \gls{ATE} F1-scores. This drop indicates that, without a \textit{None} class, the model struggled to systematically identify aspect terms. While the model became more robust during training with the \textit{None} class, its inclusion introduced ambiguity into aspect term annotation. Since aspect terms tied to the \textit{None} class did not logically serve as targets of emotions, the model may have struggled to transition between emotion and non-emotion contexts. %Despite this logical flaw, the added consistency provided by the ‘None’ class for the training dataset made it a necessary trade-off for the purposes of robust fine-tuning in ABEA.

\subsubsection{NLP-Appropriate Aspect Terms}

% definition of aspect terms
Due to the nuanced and subjective nature of emotional expression, the conceptualization of what constitutes an aspect term further added to labeling uncertainties. The guidelines therefore established a strict definition of aspect terms, emphasizing nouns, pronouns, and verbs as suitable targets while largely excluding adjectives, articles, and descriptive phrases. This restriction was aimed at making the aspect terms consistent and easier for a model to learn through predictable patterns.

% human perception
However, this computational focus diverges from the more nuanced human perception of what constitutes an aspect term. In reality, emotions are often directed at broader or more complex targets, extending beyond the central noun in a text. For example, in a Tweet such as "The decisions made by local officials have been disappointing," the emotional target includes the entire phrase "decisions made by local officials" rather than just "decisions" or "officials” on their own. Despite this discrepancy, limiting the scope of aspect terms was essential for systematic model learning, which relies on consistent input structures.

% more loose definition of aspect terms
To test the effect of a less restrictive aspect term definition, a revised version of the dataset was created with a looser annotation of emotional targets. The preliminary results (cf. Figure~\ref{fig:f1_score_config_loose_aspect_terms}) indicated that the model had significantly more difficulty extracting aspect terms, only reaching an F1-score of 46\% with the original hyperparameter configuration, compared to the F1-score of 65\% with the stricter definition. These results highlighted the need to define aspect terms in a way that balances model learnability while maintaining accurate representations of emotion targets.

\subsubsection{Inter-annotator agreement}
% in-person labeling
To achieve high \gls{IAA}, the annotation process incorporated several strategies to enhance consistency among annotators. The annotators attended in-person sessions \citep{Luo.2022, Maitama.2021}, enabling discussions to resolve ambiguities as they progressed through the dataset. %This approach, inspired by \citep{Luo.2022, Maitama.2021}, aimed to balance individual judgment with collaborative resolution of uncertainties. Annotators were encouraged to label Tweets independently but had the option to consult peers in cases of doubt, which helped mitigate potential subjective biases.
% aspect terms
The effectiveness of this annotation approach was reflected by the fact that 89\% of all aspect term annotations were included in the final dataset, based on the criteria of being labeled by at least two out of three annotators. This statistic was calculated at the character-level, i.e., the 11\% of excluded characters also contained individually annotated whitespaces around aspect term sequences, or additional words adjacent to the final aspect terms. 
% emotion labels
For the emotion labels, each identified aspect term sequence was similarly evaluated for a dominant emotion, which was identified in all but 12~cases that were individually reviewed. This high \gls{IAA}, which enabled the majority of the initial dataset to be included in the final \gls{ABEA} training dataset, underscores the value of annotator discussions. %Throughout the annotation sessions, discussions frequently emerged over both the aspect terms and their emotion classes. In most cases, different reasoning strategies led to different initial interpretations of the Tweets, which were quickly resolved after only brief conversations.

% critical discussion
From a social perspective, however, this methodological choice requires critical examination. Group dynamics, especially if pre-existing relationships or authority figures are involved, can bias annotators' objectivity. Peer conformity theory \citep{Asch.1951, Asch.1955} suggests that individuals may align with dominant voices in group settings to avoid conflict, which can result in “groupthink”, biased labels, or loss of diverse interpretations.
%In the group annotation sessions, it cannot be ruled out, that individuals were influenced in their responses. In particular, when the project organizer was present in the sessions, it may even be likely that some annotators acted in ways that avoided conflict. 
To counteract this issue, annotators were repeatedly assured that their own interpretations were valued and that no single person's opinion was inherently more correct than any other. Any residual effects of group dynamics cannot be quantified precisely and must be accepted as a baseline risk in collaborative annotation.

\subsection{Model training}

\subsubsection{Model performance plateau}
% comparison to ABSA
Using the \gls{ABEA} training dataset, the \gls{GRACE} model was fine-tuned based on the highest average validation scores in the \gls{HPO} configuration. Nevertheless, the training and validation results for \gls{ABEA} consistently showed a clear upper boundary in terms of its performance on the joint \gls{ATE} and \gls{AEC} tasks. The 10-fold cross-validation yielded an F1-score of 48.0\% for the joint task, while the gls{ATE} task alone achieved 69.1\%. By comparison, \citet{Luo.2020} reported 58.3\% for the joint tasks and 75.7\% for the gls{ATE}-only task on the \gls{ABSA} Twitter dataset from \citet{Mitchell.2013}.
On the surface, these results, i.e., 10.3 and 5.8~point drops in comparative performance, can be intuitively attributed to the increased complexities of both the new dataset and the \gls{ABEA} task. The training dataset consisted of more diverse word classes for aspect terms and was less domain-bound than conventional \gls{ABSA} training datasets (cf. Figure~\ref{fig:training_data_word_types}). In addition, the identification of emotions rather than sentiment polarities introduced more granularity. Although these changes increase model learning difficulty and can reasonably be expected to cause a drop in performance relative to the \gls{ABSA} baseline, a closer inspection of the training results provides deeper insights into the underlying causes for the model’s evident performance plateau.

% overfitting
Following the \gls{HPO}, the cumulative results revealed that the \gls{GRACE} model was not learning generalizable patterns for the two \gls{ABEA} tasks, with training performance logs indicating a strong tendency towards overfitting on the training data. This tendency was apparent from a substantial difference between training and validation performance and could be traced back to the original hyperparameters and training settings reported by \citet{Luo.2020} as optimal for the \gls{ATE} and \gls{ASC} tasks, which also suffered from significant overfitting (cf. Figure~\ref{fig:f1_score_overfitting}).
To reduce overfitting, several configurations were tested during \gls{HPO}. Most effective among these were the increased dropout rates in \gls{BERT}’s hidden layers and the attention heads. At a dropout rate of 0.3, i.e., any given node connection in its network layers with 30\% probability is dropped, the model’s training performance plateaued at F1-scores of 73\% for the \gls{ATE} task only and 61\% for the joint \gls{ATE} and \gls{AEC} tasks, respectively. While this level of dropout successfully prevented overfitting, it did not lead to a more robust performance. Instead, validation metrics significantly dropped, indicating signs of underfitting, as reflected in low performance across both training and validation data.

\begin{figure}[h]
\centering
\includegraphics[width=1\linewidth]{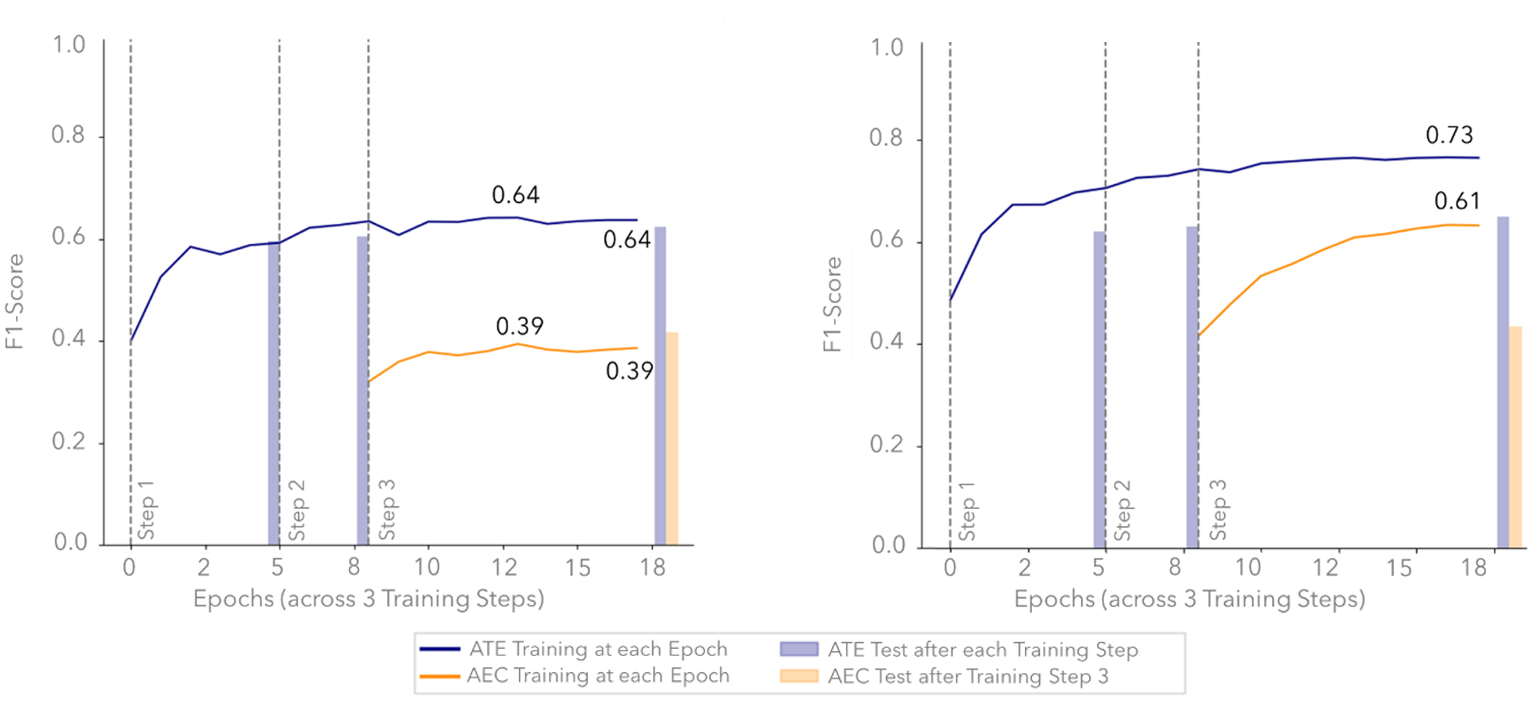}
\caption{Left: Validation metrics per epoch, Right: Training metrics per epoch during \gls{HPO} when trying to adjust for overfitting (here showing hyperparameter configuration~3)}
\label{fig:f1_score_underfitting}
\end{figure}

Consequently, the primary challenge during \gls{HPO} was to strike a balance between avoiding underfitting and preventing overfitting. The optimization process highlighted this delicate trade-off, where reducing overfitting led to notable drops in both training and validation performance. This swing from overfitting directly into underfitting was evident in the validation metrics shown in Figure~\ref{fig:f1_score_underfitting}, where performance failed to improve and instead plateaued at relatively low F1-scores of 0.64 for \gls{ATE} and 0.39 for \gls{AEC}.
%This behavior suggests that finding the optimal balance between overfitting and underfitting for the \gls{ABEA} task is particularly challenging. Unlike simpler classification tasks, \gls{ABEA} demands the learning of complex and context-specific relationships, which can easily be lost with too much regularization. 
The direct trade-off between decreased validation performance with regularization and decreased training performance without regularization implied that the \gls{GRACE} model struggled to generalize patterns from the training data, regardless of regularization.

% training dataset
In light of this evident performance plateau, a critical examination of the training dataset itself is essential Since measures to ensure label consistency were taken, the weakness is therefore likely connected to data quantity \citep{Ezen-Can.2020}. %Towards this end, research on the effect of small training datasets for \gls{BERT} fine-tuning offers conflicting arguments. 
\citet{Mosbach.2021} argue that \gls{BERT} fine-tuning instability is not caused by small training datasets themselves, but rather by the reduction of training iterations which often accompany the use of smaller datasets. However, in this study, several settings for increased training iterations did not yield significant performance improvements. 
A small dataset like the 2,621-instance \gls{ABEA} dataset likely hinders model learning for complex tasks by failing to provide a sufficiently diverse range of examples, resulting in narrow representations that do not capture the variability needed for real-world application.
This can lead to overfitting, where the model memorizes specific patterns in the training dataset, or underfitting when regularization is applied due to a lack of representative data.

% aspect terms
The impact of a too small dataset is further supported when considering \gls{ATE} and \gls{AEC} performances. Across all \gls{HPO} configurations, the performance for \gls{ATE} was considerably higher than for \gls{ATE} and \gls{AEC} combined. A likely reason for the higher \gls{ATE} performance is the amount of training examples available for aspect terms, as it includes all 5,419~terms across emotion classes. By comparison, for each emotion class, far fewer examples were available, e.g., only 282 for the class \textit{Fear}. The \gls{ATE} task therefore benefited from a more diverse training set, even with the added complexity of more varied word types compared to prior \gls{ABSA} datasets. 
% emotion classes
In contrast, the number of training examples per emotion class was much lower, especially for the classes \textit{Fear} and \textit{Sadness}.
Additionally, many words were unique to a single emotion class. While this may be expected for a small dataset, it increases the risk of the model learning specific words rather than generalizable patterns. In future work, data augmentation techniques should be considered to expand small training datasets without requiring further time-consuming annotation. This would be a viable option to mitigate overfitting and help \gls{NLP} models to generalize learned patterns.

% The finding that the original HPO configuration by \citep{Luo.2020} already caused extensive overfitting, which was not reported by the authors, gave a somewhat misleading impression of the model’s capabilities. Had Luo et al (2020) transparently reported not just validation but also training metrics, it may have been possible to adjust the methodology accordingly from the outset, or by extension, potentially not selecting the GRACE model as the target architecture for this thesis. The detailed methodology, results and discussion sections of this thesis aim to add transparency to the research in this field of research. 

\FloatBarrier

\subsubsection{Hyperparameter settings}

% general impact of hyperparameters
Throughout the \gls{HPO} process, 16~hyperparameter categories were adjusted. Some settings, such as weight decay, layer reduction, and layer freezing, were aimed at mitigating overfitting. Other modifications were intended to improve learning capacity, such as incorporating more complex classifier architectures or introducing earlier split between the \gls{ATE} and \gls{AEC} branches. Overall, the average model performance was relatively insensitive to changes in hyperparameters, with only minor fluctuations across the different \gls{HPO} configurations. This suggests that insufficient dataset size, rather than hyperparameter settings, may be the primary limiting factor in model performance. Nonetheless, some hyperparameters adjustments did yield noticeable effects on model outcomes.

% classification heads
One of the most influential changes was the adjustment of the classification heads. Introducing more complex classification head architectures, specifically DenseNet-inspired classifiers, led to slight performance gains in the \gls{AEC} task. However, this improvement came with a significant trade-off for the \gls{ATE} task, particularly on the SemEval validation datasets, ultimately reducing overall performance substantially. 
The increased complexity of the classification heads likely exceeded what the dataset size could support, causing more severe overfitting on the training data. %With larger training datasets, it may be feasible to test alternative classifiers. 
To balance training and validation performance, the final model retained a linear layer as classification head.

% VAT
\gls{VAT} was originally included in the training process of \gls{GRACE} during the second training step. However, \gls{HPO} results indicated that using \gls{VAT} had no positive impact on model performance. One plausible explanation is that the \gls{ABEA} training data is complex enough without additional adversarial examples, making \gls{VAT} unnecessary or even detrimental. As a result, \gls{VAT} was removed from the final model configuration to streamline the training process. For an augmented dataset, however, \gls{VAT} should be re-considered and may achieve its intended effect of increasing model robustness.

% GHL
Various \gls{GHL} configurations were tested, including its application at different training steps, adjustments to the number of bins (used to group and assign weights to training examples based on their difficulty), and modifications to momentum values. While increasing the number of bins from 24 to 32 yielded the highest average performance by allowing finer differentiation of training examples based on difficulty, none of the \gls{GHL} adaptations resulted in a significant improvement over previously achieved performances.
As a result, the original \gls{GHL} settings were retained. However, the underlying assumption of the \gls{GHL} mechanism should be critically evaluated in and of itself. \gls{GHL} assumes that the most frequent labels are also the easiest to learn. However, this may not apply to emotion classification, where the class distribution does not necessarily correlate with learning difficulty. In the new \gls{ABEA} dataset, it is not at all self-evident that the most frequent classes, \textit{Happiness} and \textit{None}, are easier to learn compared to the least frequent classes \textit{Fear} and \textit{Sadness}. In contrast, for the original \gls{ABSA} dataset \citep{Mitchell.2013}, \gls{GHL} is far more applicable since 70\% of all examples belong to the \textit{Neutral} sentiment class. %The \gls{GHL} mechanism itself may therefore not be optimally suitable for \gls{ABEA} training compared to \gls{ABSA} training.

% shared & decoded layers
Adjustments to the number of shared layers between the \gls{ATE} and \gls{AEC} branches, as well as the number of decoder layers in the \gls{AEC} branch, led to minor improvements in average performance. Originally, the \gls{GRACE} model used nine shared layers, with only two decoder layers in the \gls{AEC} branch. Interestingly, experiments showed that reducing the number of shared layers to five and accordingly increasing the \gls{AEC} branch’s decoder layers to six improved model performance. This suggests that the model was already learning the \gls{ATE} task satisfactorily, allowing the additional dedicated layers in the \gls{AEC} branch to enhance its ability to capture task-specific patterns.

% layer freezing
Layer freezing was tested as a regularization strategy to mitigate overfitting. Configurations included freezing the bottom three, five, and nine layers of BERT-base’s twelve overall layers. Surprisingly, freezing the bottom three and five layers had minimal impact on the model’s performance, either positively or negatively, with the averaged performance remaining at 49.4\% and 49.5\%, respectively. Freezing the bottom nine layers, however, did lead to a noticeable performance drop to an average of 45.1\%. This finding further reinforced the notion that hyperparameter tuning had limited effect and suggested the dataset constraints, rather than the model structure, as the primary bottleneck for performance gains.

% % HPO process
% In consideration of the \gls{HPO} process as a whole, this work underscores the need for more efficient and feasible methods, especially for complex models like \gls{GRACE} with numerous hyperparameter combinations across several training steps. The manual adjustments in this study were necessary to implement extensive validation in between iterations but were time-consuming and limited the scope of combinations tested. Future work could explore automated \gls{HPO} techniques and how to adjust the fine-tuning process to incorporate validation on external datasets. %Additionally, an evaluation of the model’s capabilities on separate data should be carried out.

\section{Conclusion}

This paper addresses existing research gaps in the field of \acrlong{ABEA} by creating a consistent training dataset (\textbf{RQ1}) as well as adapting and fine-tuning the \gls{GRACE} model \citep{Luo.2020} for the joint extraction of the sub-tasks \gls{ATE} and \gls{AEC} (\textbf{RQ2}).

To address the training dataset bottleneck for \gls{ABEA}, a Twitter dataset was labeled using group annotation and majority voting to ensure label consistency. In particular, the introduction of a \textit{None} emotion class and a tightening of aspect term definitions addressed the conceptual ambiguities inherent to emotion-classification and \gls{NLP} at the token-level, leading to improved annotation coherence. % Based on a two-step character-level majority voting system, the aspect terms were first finalized with an \gls{IAA} of 89\%, followed by the determination of the final emotion labels, where only 12~non-majority conflicts required resolution.
The final dataset comprised 2,621~Tweets.

To repurpose the \gls{ABSA} model \gls{GRACE} \citep{Luo.2020} for \gls{ABEA}, extensive \gls{HPO} was conducted. The most impactful hyperparameter changes on model performance included reductions in batch size, a warm-up method adjustment, a reduction in the number of shared layers between the \gls{ATE} and \gls{AEC} branches, an increase of \gls{AEC} branch layers, extending training epochs, and omitting \gls{VAT} in the second training step. A more complex classification layer negatively impacted model performance. The fine-tuned \gls{ABEA} model plateaued at F1-scores of 70.1\% for \gls{ATE} and 46.9\% for joint \gls{ATE} and \gls{AEC}.

The lower performance of EmoGRACE for \gls{ABEA} compared to the reported performance of \gls{GRACE} for \gls{ABSA} can be largely attributed to three factors. First, the shift from sentiment polarity to more complex emotion classification. Second, the increased variance in types of words for aspect terms, extending beyond just nouns. Third, the size of the training dataset proved insufficient to allow for generalizable model learning at the high level of task complexity, indicating that data augmentation may be a necessary pre-requisite for continued \gls{ABEA} research.
% future work
Future work may consider the use of generative foundation models, such as \gls{GPT} models \citep{Radford.2018}, which have shown strong capabilities in handling a variety of \gls{NLP} tasks and may even outperform fine-tuned task-specific \gls{BERT} models in certain applications \citep{Zhong.2023}.
 
% paragraph about application
% This model was designed for processing disaster-related social media posts, with the end goal of identifying fine-grained emotional patterns to inform disaster response. 
The effectiveness of the presented \gls{ABEA} method in capturing public emotions should be evaluated through a targeted, domain-specific case study, such as analyzing posts during a specific natural disaster. The model could be applied to the social media-based analysis of other highly emotional social topics, such as public health crises \citep{Kogan.2021}, climate change \citep{Li.2023a}, urban events \citep{Honzak.2024} or political discourses \citep{Kovacs.2021}, providing insights into collective emotions.

\backmatter

% \bmhead{Supplementary information}

% If your article has accompanying supplementary file/s please state so here. 

% Authors reporting data from electrophoretic gels and blots should supply the full unprocessed scans for key as part of their Supplementary information. This may be requested by the editorial team/s if it is missing.

% Please refer to Journal-level guidance for any specific requirements.

\bmhead{Acknowledgements}

This project has received funding from the European Commission - European Union under HORIZON EUROPE (HORIZON Research and Innovation Actions) under grant agreement 101093003 (HORIZON-CL4-2022-DATA-01-01). Views and opinions expressed are however those of the author(s) only and do not necessarily reflect those of the European Union - European Commission. Neither the European Commission nor the European Union can be held responsible for them.

This paper also received funding through the Austrian Research Promotion Agency (FFG) and the DigitalInnovationLayer project (FO999898902).

We would like to thank
David Hanny and Shaily Gandhi (IT:U) for their feedback.

\bmhead{Author contributions}
Conceptualization, C.Z., B.R.; 
methodology, C.Z.; 
software, C.Z.; 
validation, C.Z.; 
formal analysis, C.Z.; 
investigation, C.Z.; 
resources, B.R.; 
data curation, C.Z., S.S.; 
writing---original draft preparation, C.Z., S.S.; 
writing---review and editing, C.Z., S.S.; 
visualization, C.Z.; 
supervision, B.R.; 
project administration, B.R.; 
funding acquisition, B.R. \\ 
All authors have read and agreed to the published version of the manuscript.

\begingroup
\renewcommand{\normalsize}{\small} % Adjust font globally for this scope
\setstretch{0.8} % Adjust line spacing (default is 1.0, reduce for tighter spacing)
\bibliography{bibliography}
\endgroup

%% if required, the content of .bbl file can be included here once bbl is generated
%%\input sn-article.bbl

\newpage
\begin{appendices}

\section{}\label{secA1}

\begin{center}
\begin{table}[ht]
\centering
\caption{Training datasets found in relation to the search terms “aspect-based emotion analysis”, “aspect-based emotion dataset”, or “fine-grained emotion dataset”. The dataset used in this paper is featured in the last row.}
\label{tab:abea_training_datasets}
\begin{tabular}{m{1.25cm}m{1cm}m{3.5cm}m{1cm}m{4cm}}
\toprule
\textbf{Dataset} & \textbf{Public} & \textbf{Data Source} & \textbf{Size} & \textbf{Comment} \\ \midrule
\cite{DeBruyne.2022}& Yes& Adidas Instagram account (images and comments) & 4,900 & Multimodal dataset\\
\cite{Garcia-Mendez.2023}& No & Twitter & 5,000 & Dataset and model serve the understanding of financial/stock market risk and opportunity\\
\cite{Maitama.2021} & No & SemEval Restaurant dataset \citep{Pontiki.2014} & 3,850 & Multilingual, mostly Indonesian language; After \gls{IAA} the authors removed emotions ‘anger’, ‘fear’ and ‘disgust’ due to small cohorts \\
\cite{Suciati.2020} & No & Indonesian restaurant review platform ‘PergiKuliner’ & 14,103& Multilingual (English, Dutch, German, French); designed for a business platform\\
\cite{DeGeyndt.2022} & No & Bol.com, Trustpilot, TripAdvisor, and proprietary data from project partners & 1,000+& Hungarian language \\
\cite{Uveges.2022} & No & Hungarian speeches from 2014 - 2018 & 1,008& Aspects are not labeled per text, but considered as overall categories for which emotion scores are identified; emotions are based on COVID-19 death rates\\
\cite{DeviT..2021} & No & Twitter & unclear & Aspects are not labeled per text, but considered as overall categories for which emotion scores are identified; emotions are based on COVID-19 death rates\\ \midrule
 & Yes & Twitter & 2,621 & English dataset with complex aspect term annotation, four emotions and a \textit{None} class. \\
\bottomrule
\end{tabular}
\end{table}
\centering
\end{center}

\begin{figure}[h]
\centering
\includegraphics[width=0.8\linewidth]{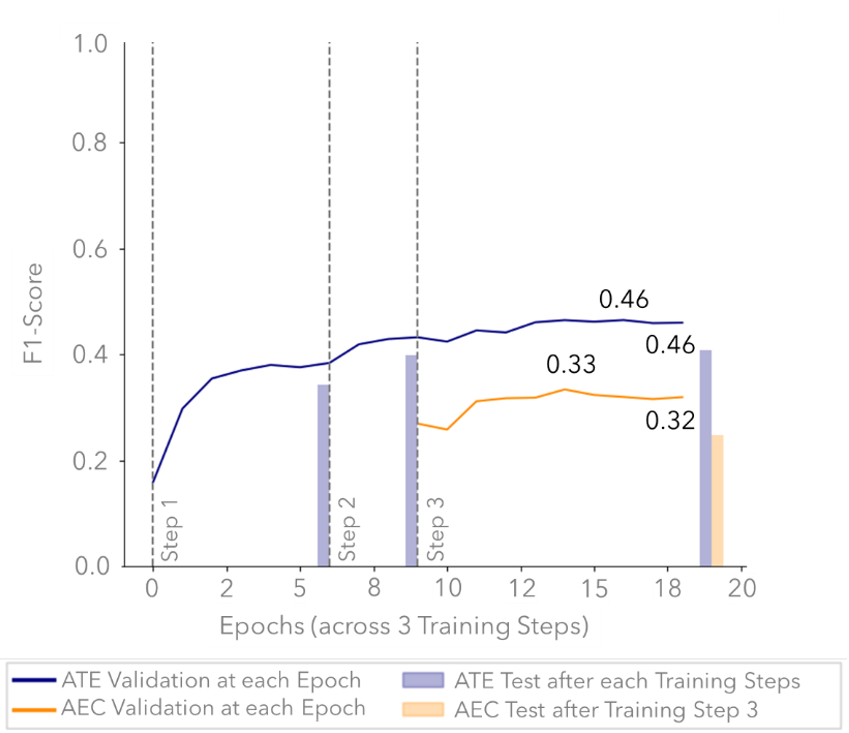}
\caption{Training metrics on an adapted training dataset with more loosely defined aspect terms.}
\label{fig:f1_score_config_loose_aspect_terms}
\end{figure}

\begin{figure}[h]
\centering
\includegraphics[width=1\linewidth]{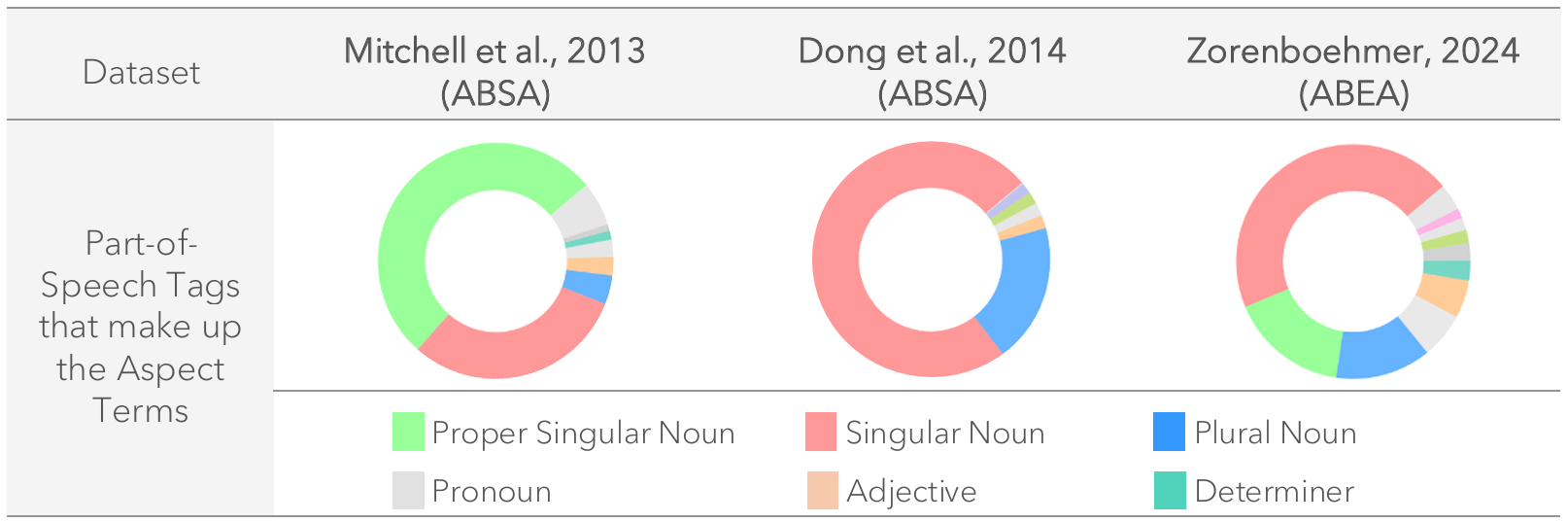}
\caption{Direct comparison of the type of words labeled as aspect terms among ABSA and ABEA datasets.}
\label{fig:training_data_word_types}
\end{figure}

\begin{figure}[h]
\centering
\includegraphics[width=1\linewidth]{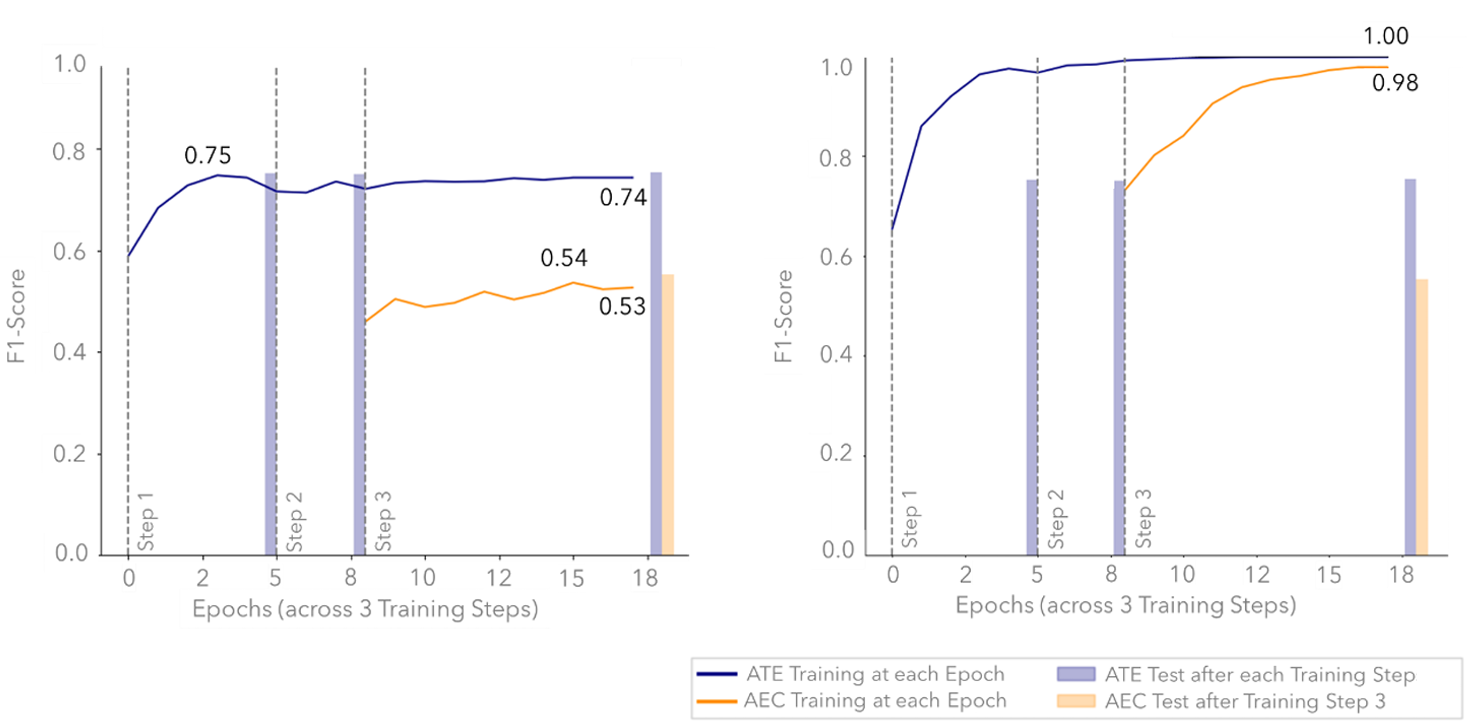}
\caption{Left: Validation metrics per epoch, Right: Training metrics per epoch using the original \gls{ABSA} dataset with the original hyperparameters and training settings as reported by \citet{Luo.2020}}
\label{fig:f1_score_overfitting}
\end{figure}

\FloatBarrier
\end{appendices}

\end{document}